\documentclass{article}

\usepackage{PRIMEarxiv}

\usepackage{xcolor}
\usepackage{makecell}
\usepackage{amsmath,amsfonts,amssymb,amsthm,mathtools}
\usepackage{booktabs}       
\usepackage{amsfonts}       
\usepackage{nicefrac}       
\usepackage{microtype}      
\usepackage{lipsum,bm}
\usepackage{graphicx}
\usepackage{ragged2e}
\usepackage{multirow}
\usepackage{array}
\usepackage{threeparttable}
\usepackage{pifont}
\usepackage{enumitem}
\usepackage{lineno}
\def\etal{\textit{et al. }}

\newtheorem{definition}{Definition}
\newtheorem{theorem}{Theorem}

\title{Circular Quasiconformal Deturbulence: Geometry-Based Restoration from Multiple Turbulent Frames}

\author{
  Chu Chen \\
  Department of Mathematics, City University of Hong Kong \\
  Hong Kong Centre for Cerebro-Cardiovascular Health Engineering \\
  Hong Kong\\
  \texttt{{chuchen4-c}@my.cityu.edu.hk} \\
    \And
  Han Zhang\\
  Department of Mathematics, City University of Hong Kong \\
  Hong Kong Centre for Cerebro-Cardiovascular Health Engineering \\
  Hong Kong\\
  \texttt{hzhang863-c@my.cityu.edu.hk} \\
   \AND
  Lok Ming Lui \thanks{Corresponding Author.}\\
   Department of Mathematics, Chinese University of Hong Kong \\
   Hong Kong\\
   \texttt{lmlui@math.cuhk.edu.hk} \\
}

\begin{document}
\maketitle
\begin{abstract}
Imaging through inhomogeneous media often results in severe distortions, posing significant challenges to downstream image-processing tasks. The lack of clean paired images makes supervised learning impractical, motivating unsupervised restoration approaches. In this work, we propose the Circular Quasi-Conformal Deturbulence (CQCD) framework, an unsupervised approach that reconstructs distortion-free images from multiple frames using a circular architecture. The framework minimizes reconstruction errors by jointly estimating forward and backward transformations between distorted observations and the restored image. A key advancement of CQCD is the integration of computational quasi-conformal geometry, which encourages bijective non-rigid deformations and improves the well-posedness of both forward and inverse mappings for cycle consistency. The deformation field is further regularized to preserve structural coherence and reduce non-physical artifacts such as folding or tearing. Additionally, tight-frame blocks are employed to effectively encode distortion-sensitive features, enhancing the precision of the restoration process. To assess the effectiveness of the proposed framework, extensive evaluations are conducted on synthetic and real-world image datasets.   Experimental findings indicate that CQCD not only surpasses existing state-of-the-art deturbulence techniques in restoration quality but also achieves highly accurate deformation field estimation.
\end{abstract}

\keywords{Image Restoration \and Turbulence Removal \and Quasi-Conformal Geometry \and Circular Model \and Tight Frame \and Water turbulence}
\section{Introduction}

Image restoration is a fundamental problem in computer vision that seeks to recover a clear, high-quality image from its degraded version. This task is crucial in numerous real-world applications, including medical imaging, surveillance, remote sensing, and digital photography, where accurate interpretation and analysis rely on high-fidelity image reconstruction. The ability to effectively restore images from noise, blur, and geometric distortions significantly enhances the performance of downstream tasks such as object recognition, segmentation, and classification.

A major challenge in image restoration arises from the lack of high-quality, paired label images for training supervised models. Supervised learning typically requires pairs of distorted and corresponding clear images, which are often difficult or even impossible to obtain in practice. For instance, in scenarios involving atmospheric turbulence, optical aberrations, or underwater imaging, capturing both the degraded and ground-truth images under identical conditions is infeasible. To address this limitation, existing approaches attempt to synthesize distorted images from available clear images using mathematical models or learn distortions from sequential frames in videos. While effective in certain cases, these methods often struggle to generalize across different distortion types and may introduce artifacts due to imperfect distortion modeling.

In this work, we introduce a novel unsupervised learning framework for image restoration, named the Circular Quasi-Conformal Deturbulence (CQCD) model. Our approach is designed to recover a distortion-free scene given sequences suffered from complex geometric distortions, which typically arise from inhomogeneous media such as turbulent air, water, or other refractive environments. Unlike methods that rely on explicit supervision or approximate distortion priors, our model learns to restore images by leveraging a circular self-consistency constraint across input sequences. The core idea is to estimate a set of deformation mappings that warp the distorted images into a geometrically corrected form, followed by a deblurring process to enhance clarity.

The CQCD model operates in a three-stage cycle. First, a sequence of distorted images undergoes feature extraction via a tight-frame transformation, which enhances structural details while reducing noise. These features are then fed into a mapping estimator, a neural network that predicts the deformation fields necessary to geometrically correct the images. The warped images, now free of geometric distortions, still retain residual blur due to resampling and the original degradation. To address this, a blur removal module is applied, producing a sharp, restored image. The final stage promotes cycle consistency by verifying whether the estimated inverse mappings can reintroduce the original distortions. Specifically, by applying the inverse deformation to the restored image, we reconstruct an approximation of the input distorted images. If these reconstructed distortions align with the original inputs, the estimated mappings are deemed accurate. This encourages the learned transformations to faithfully model the true geometric distortions.

The circular architecture offers several advantages. First, it promotes temporal and spatial consistency, helping restored images maintain structural integrity across multiple frames. Second, it acts as an implicit validation mechanism, reducing the accumulation of incorrect estimations. If the learned deformation mappings are inaccurate, the inverse transformations will fail to reconstruct the original distortions, providing a built-in feedback mechanism for self-correction. This iterative refinement leads to progressively improved restoration performance.

Since the restoration process involves both forward and inverse mappings, it is crucial to encourage these transformations to be invertible and stable. In our model, we incorporate quasi-conformal geometry theory as guidance for homeomorphic deformations. A key mathematical tool in this process is the Beltrami coefficient, which quantifies the local distortion of a mapping. By regularizing the Beltrami coefficient within a controlled range, we encourage the estimated deformation fields to remain smooth and bijective, reducing folding or overlapping artifacts in practice. This promotes preservation of topological structures while effectively reversing distortions.

To assess the effectiveness of CQCD, we conduct experiments on synthetic and real-world images affected by various geometric distortions. Comparisons with unsupervised methods across different frame counts underscore the robustness of CQCD in correcting such distortions. Tests on real distorted images further demonstrate the model’s practical effectiveness in challenging scenarios. Additionally, we validate the accuracy of the estimated deformation fields by comparing them with ground-truth fields used to simulate turbulence. Ablation studies on the quasi-conformal (QC) regularization weighting and the inclusion of the tight-frame (TF) block provide insights into optimal QC weighting and highlight the significance of TF-based encoding.

In summary, our contributions are as follows:
\begin{enumerate}
    \item We propose a circular turbulence removal framework based on deformation mappings that utilizes multiple observations. The circular architecture promotes accurate propagation of the restored image and mitigates the accumulation of estimation errors.
    \item We incorporate computational quasi-conformal theory to guide the generated mappings toward homeomorphic and invertible behavior. This is essential for the circular architecture, which requires both the forward deformation mapping and its inverse for restoration.
    \item We introduce tight-frame encoding to enhance feature extraction, enabling the CNN to directly capture orientational deformation patterns. This facilitates more accurate deformation estimation and improved blur correction.
\end{enumerate}
\section{Related Work} In this section, we systematically review the literature that is closely aligned with our research. We focus on two major areas: computational quasiconformal geometry and image restoration, highlighting key contributions and methodologies.

\subsection{Computational quasiconformal geometry}
Computational quasiconformal geometry has emerged as a powerful mathematical framework for analyzing and manipulating geometric distortions across various imaging tasks. It provides a systematic approach to control distortions under mappings, as demonstrated in foundational works such as \cite{lam2014landmark, lui2014teichmuller}. Specifically, conformal mappings, a subset of quasiconformal mappings, have proven effective in multiple geometry processing applications, ranging from texture mapping to surface parameterization \cite{levy2002least, gu2004genus, gu2003global}.

The Beltrami coefficient is extensively used to quantify local geometric distortions within mappings. By adjusting the Beltrami coefficients, researchers can effectively control geometric properties, enabling the development of surface parameterization techniques aimed at minimizing conformality distortion \cite{choi2016spherical, choi2020free}.

For image registration, Lam et al.~\cite{lam2014landmark} integrate landmark correspondences with intensity information through quasi-conformal mapping, while Lui et al.~\cite{lui2014teichmuller} employ Teichmüller mappings to achieve similar goals. In surface matching, \cite{choi2015fast} leverages polar projection and reparameterization for genus-0 geometries. More recently, Zhang et al.~\cite{zhang2025quasi} introduced deformation convolution on manifolds through quasi-conformal mapping, advancing techniques in geometric reparametrization. Quasiconformal mappings have also been incorporated into segmentation tasks by warping template masks, as seen in \cite{zhang2021topology, zhang2024learning}, with further integration of user inputs for interactive refinement proposed in \cite{zhang2025qis}. Additionally, uncertainty modeling in deformation analysis has been explored in \cite{zhang2022nondeterministic, zhang2022new}, focusing on medical image analysis and disease assessment.

\subsection{Image Restoration}
Generative Adversarial Networks (GANs) have gained prominence in the realm of image restoration, particularly in restoring degraded images to their original, high-quality states \cite{kupyn2018deblurgan, kupyn2019deblurgan, zhu2017unpaired, isola2017image}. By training a generator to reconstruct high-quality images from degraded inputs, these models effectively produce visually realistic and perceptually accurate results.

In the domain of image deturbulence, Lau et al. \cite{lau2019restoration} utilize robust principal component analysis (RPCA) alongside quasiconformal mappings to mitigate atmospheric turbulence distortions. Thepa et al. \cite{thapa2020dynamic} present a neural network-based approach to reconstruct dynamic fluid surfaces from monocular and stereo images. A more recent approach by Anantrasirichai \cite{anantrasirichai2023atmospheric} leverages complex-valued convolutions to enhance atmospheric image restoration.

Lau et al. \cite{lau2020atfacegan} further extend GAN applications by introducing a specialized model for facial image restoration under turbulent conditions. Zhang et al. \cite{zhang2025deformation} employ homeomorphic mapping for geometric restoration of distorted data, while Li et al. \cite{li2018learning} focus on refractive distortions, proposing a GAN-based model to correct water surface distortions using single images. Additionally, Jiang et al. \cite{jiang2023nert} adopt a general implicit neural representation for unsupervised turbulence removal. Rai et al. \cite{rai2022removing} incorporate a channel attention mechanism, adapted from \cite{hu2018squeeze}, into their GAN model to focus on critical features during the restoration process. Pan \etal~\cite{pan2021tsan} introduced a two-stream attention network for synthesized view quality enhancement in 3D-HEVC, fusing global context and local detail to improve restoration. This was followed by a multi-module cascade network for mobile image enhancement~\cite{pan2021miegan}. Lin et al. \cite{lin2023unsupervised} introduced a self-collaboration strategy with parallel generative adversarial branches for unsupervised real-world denoising, demonstrating that iterative replacement of the denoiser within a noise extraction module can boost performance without increasing inference complexity. This idea was further extended in \cite{lin2025re} to general image restoration tasks via a prompt learning mechanism.

\subsection{Circular Model}
Our work is strongly inspired by the principle of cycle-consistency, which has been widely used to improve learning in ill-posed problems without paired data. A first example is CycleGAN~\cite{zhu2017unpaired}, which enforces consistency between forward and backward mappings to enhance image generation. Kim \etal~\cite{kim2017learning} applied circular consistency for style transfer across domains, while Liu \etal~\cite{liu2017unsupervised} further regularized unsupervised image-to-image translation by incorporating cycle-reconstruction streams. In geometric vision tasks such as correspondence and deformation estimation, cycle constraints help reduce ambiguity in transformation learning~\cite{zhou2016learning,wang2019learning}.

The proposed circular architecture is closely related to cycle-consistent learning paradigms, as it enforces reconstruction consistency through forward–backward transformations. Unlike prior approaches that learn unconstrained mappings between image domains, our method integrates this principle within a geometry-aware quasi-conformal deformation model. The quasi-conformal constraint enforces the bijectivity of the deformation mapping and its inverse, which is essential for maintaining a valid cycle-consistency loop. Without invertibility, the circular process would break, leading to error accumulation as additional frames are incorporated, as reflected by the experiment in Section~\ref{sec:qcregularization}. By coupling cycle-consistency with explicit geometric regularization, our framework is particularly well-suited for distortion-driven image restoration.

\section{Mathematical Background}

\subsection{Formulation of Image Distortion}





\begin{figure}
    \centering
    \includegraphics[width=0.5\linewidth]{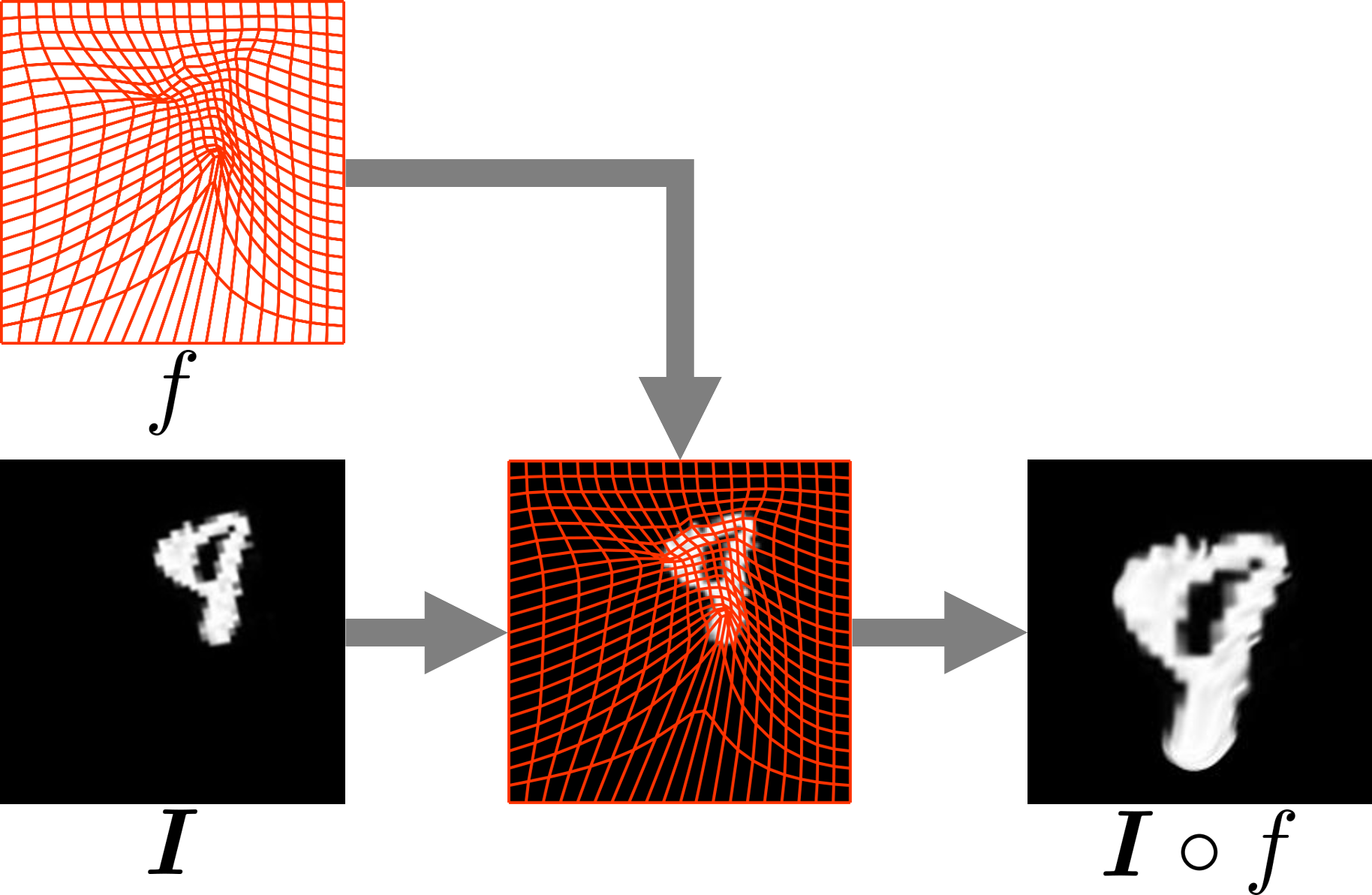}
    \caption{Visual illustration of obtaining the composited image $\boldsymbol{I}\circ f$ from the input image $\boldsymbol{I}$ and mapping $f$.}
    \label{fig:image_map_composition}
\end{figure}

A distorted video of a scene can be viewed as a sequence of frames, each affected by varying levels of degradation due to dynamic turbulence. As time progresses, each frame experiences different degrees of blurring, geometric distortion, and noise. Let \(\{\tilde{I}_t\}_{t=1}^T\) represent a sequence of distorted images over time, where \(t\) denotes the time index, and \(T\) is the total number of frames in the sequence. Each distorted frame \(\tilde{I}_t(x, y)\) at time \(t\) can be described mathematically as:

\begin{equation}
\tilde{I}_t(x, y) = (B_t * I_t \circ f_t)(x, y) + n_t(x, y),
\label{eq:distort_formula}
\end{equation}
where \(I_t(x, y)\) is the original image at time \(t\), \(f_t: \mathbb{R}^2 \to \mathbb{R}^2\) is the geometric distortion function, \(B_t(x, y)\) is the blurring kernel, and \(n_t(x, y)\) represents the additive noise. Here, $*$ denotes the convolution operation, and $\circ$ denotes composition. Compositing the image $\boldsymbol{I}$ with a mapping $f$ effectively reparameterizes the image, which can also be interpreted as a spatial transformation. A visual illustration of this operation is shown in Figure~\ref{fig:image_map_composition}. This formulation captures the combined impact of turbulence-induced degradation across the video sequence.

\subsection{Quasi-Conformal Theory}

\begin{definition}(Quasi-conformal map).
A quasi-conformal map $f: \mathbb{C} \rightarrow \mathbb{C}$ is a mapping that satisfying the Beltrami equation
\begin{equation}
\frac{\partial f}{\partial \bar{z}}=\mu(z) \frac{\partial f}{\partial z}
\label{eq:beleq}
\end{equation}
where $\mu(z)$, the Beltrami coefficient (BC), is a complex-valued function with the constraint $\|\mu\|_{\infty} < 1$, and $\frac{\partial f}{\partial z}$ is non-zero almost everywhere. The complex partial derivatives are defined as:
\begin{equation}
\begin{aligned}
\frac{\partial f}{\partial z}=\frac{1}{2}\left(\frac{\partial f}{\partial x}-i \frac{\partial f}{\partial y}\right)
\frac{\partial f}{\partial \bar{z}}=\frac{1}{2}\left(\frac{\partial f}{\partial x}+i \frac{\partial f}{\partial y}\right)    
\end{aligned}
\end{equation}
\end{definition}
\noindent The Beltrami coefficient $\mu$ serves as a quantitative indicator of how far the map $f$ deviates from being conformal. Notably, at any point $p$ where $\mu(p) = 0$, the map $f$ behaves as a conformal map in a neighborhood around $p$, and in such cases, the Beltrami equation \eqref{eq:beleq} simplifies to the classical Cauchy-Riemann equations.
\cite{chan2024classification,choi2015fast,lui2013texture}.

\begin{figure}
    \centering
    \includegraphics[width=0.4\textwidth]{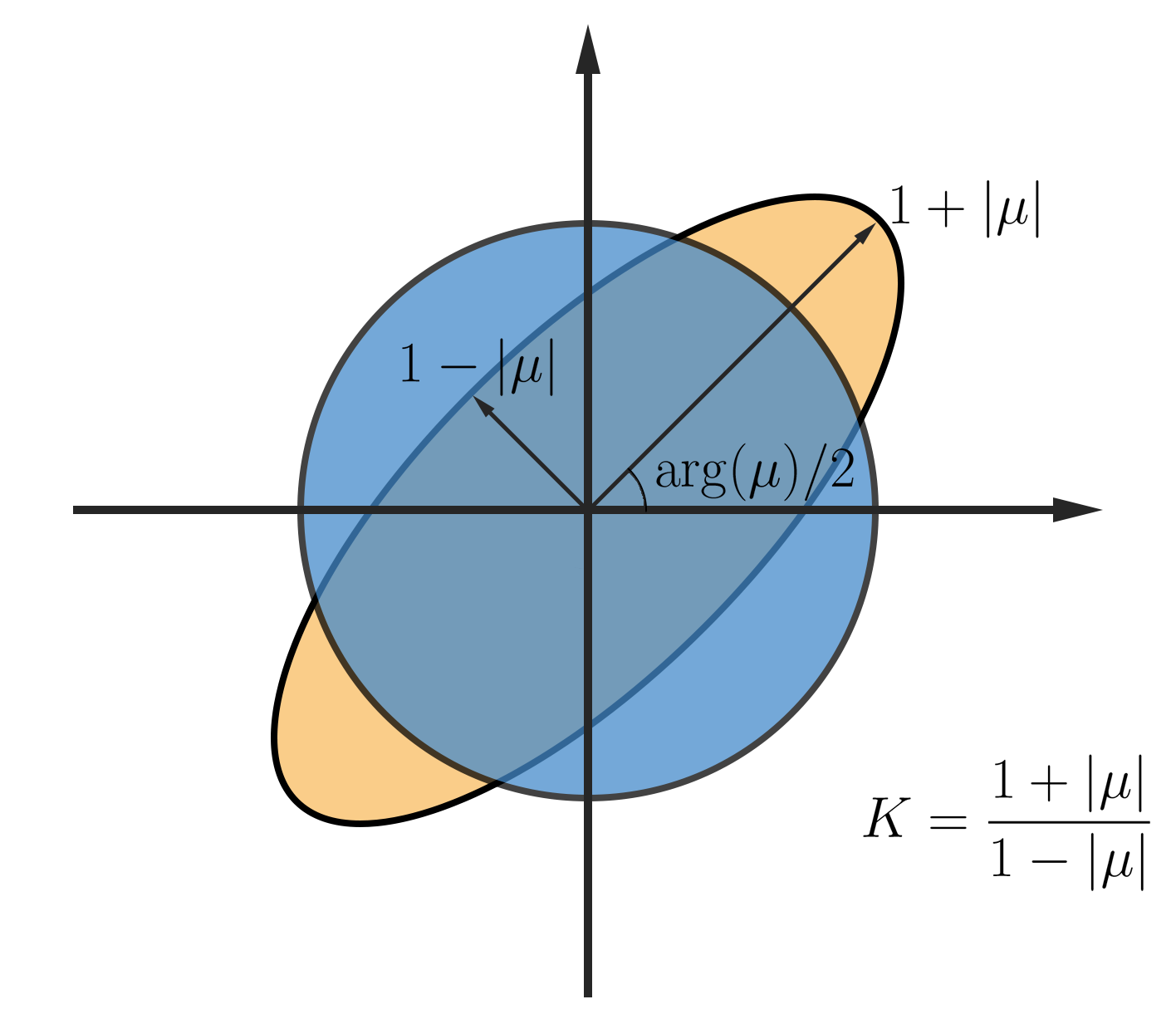}
    \caption{Illustration of how the Beltrami coefficient measures the conformality distortion of a quasi-conformal map}
    \label{fig:qcmap}
\end{figure}

In a local neighborhood around a point $p$, the map $f$ can be approximated as:
\begin{equation}
\begin{aligned}
f(z) &=f(p)+f_{z}(p) z+f_{\bar{z}}(p) \bar{z} \\
&=f(p)+f_{z}(p)(z+\mu(p) \bar{z}).
\end{aligned}
\end{equation}
Here, $f(p)$ represents a translation, and $f_z(p)$ denotes a dilation, both of which are conformal transformations. The non-conformality is solely attributed to the term $D(z) = z + \mu(p) \bar{z}$. Hence, the Beltrami coefficient $\mu$ effectively encodes the degree of conformality distortion.
The maximal quasi-conformal dilation of $f$ is given by
\begin{equation}
K=\frac{1+\|\mu\|_{\infty}}{1-\|\mu\|_{\infty}}.
\end{equation}
Figure \ref{fig:qcmap} illustrates the geometry of a quasi-conformal map.

A significant connection between a map and its Beltrami coefficient lies in the homeomorphism property on bounded domains with boundary conditions. This is formalized in the following theorem:

\begin{theorem}[Bijectivity on a bounded image domain]\label{thm:qcbc}
Let $\Omega\subset\mathbb C$ be a bounded Jordan domain, and
let\[
f\in C(\overline{\Omega})\cap C^{1}(\Omega).
\]
Assume:

\begin{enumerate}
\item $f_{z}(z)\neq 0$ for every $z\in\Omega$, and
\[
\mu(z):=\frac{f_{\bar z}(z)}{f_{z}(z)}
\quad\text{satisfies}\quad
\|\mu\|_{L^\infty(\Omega)}<1;
\]
\item the boundary map
\[
f|_{\partial\Omega}:\partial\Omega\to\Gamma
\]
is an orientation-preserving homeomorphism onto a Jordan curve $\Gamma$.
\end{enumerate}

Let $D:=\operatorname{int}\Gamma$ be the bounded component of $\mathbb C\setminus\Gamma$.
Then
\[
f:\overline{\Omega}\to\overline{D}
\]
is a homeomorphism. In particular, $f$ is bijective from $\overline{\Omega}$ onto $\overline D$, and
$f|_{\Omega}:\Omega\to D$ is bijective.
\end{theorem}

\begin{proof}
Since $\mu=f_{\bar z}/f_z$ and $\|\mu\|_\infty<1$, for every $z\in\Omega$,
\[
J_f(z)=|f_z(z)|^2-|f_{\bar z}(z)|^2
      =|f_z(z)|^2\bigl(1-|\mu(z)|^2\bigr)>0.
\]
Hence $Df(z)$ is invertible at each $z\in\Omega$, so by the inverse function theorem
$f|_\Omega$ is a local $C^1$-diffeomorphism, in particular a local homeomorphism and an open map.

Now compare Brouwer degrees using the boundary map.
Because $f|_{\partial\Omega}$ is an orientation-preserving homeomorphism onto $\Gamma$, for $y\notin\Gamma$,
\[
\deg(f,\Omega,y)=
\begin{cases}
1,& y\in D,\\
0,& y\in \mathbb C\setminus \overline D.
\end{cases}
\]
Fix $y\in \mathbb C\setminus \overline D$. If $y=f(x)$ for some $x\in\Omega$, then each preimage contributes
positive local degree (since $J_f>0$), so $\deg(f,\Omega,y)>0$, contradicting $\deg(f,\Omega,y)=0$. Therefore,
\[
f(\Omega)\cap (\mathbb C\setminus \overline D)=\varnothing,
\]
i.e. no interior point maps to the exterior of \(\Gamma\). 
Since \(f|_\Omega\) is a local diffeomorphism, it is an open map. Thus if some \(x_0\in\Omega\) satisfied \(f(x_0)\in\Gamma\), then \(f(\Omega)\) would contain an open neighborhood of \(f(x_0)\). Every open neighborhood of a point of \(\Gamma\) meets \(\mathbb C\setminus\overline D\), contradicting the previous conclusion. Hence
\[
f(\Omega)\cap\Gamma=\varnothing.
\]

\noindent We can conclude that
\[
f(\Omega)\subset D.
\]
Now, we define that $g:=f|_\Omega:\Omega\to D.$ We claim that $g$ is proper. Let $K\subset D$ be compact. Then $g^{-1}(K)=f^{-1}(K)\cap\Omega$. Since $f$ is continuous on $\overline\Omega$, $f^{-1}(K)$ is closed in $\overline\Omega$. Thus, it is compact (because $\overline\Omega$ is compact).
Also
\[
f^{-1}(K)\cap\partial\Omega=\varnothing,
\]
because $f(\partial\Omega)=\Gamma$ and $K\cap\Gamma=\varnothing$.
Hence $g^{-1}(K)=f^{-1}(K)$ is compact in $\Omega$.
So $g$ is proper.

Thus, $g$ is a proper local homeomorphism.
A proper local homeomorphism into a Hausdorff space is closed, and a local homeomorphism is open. Hence $g(\Omega)$ is both open and closed in the connected set $D$. Since $g(\Omega)\neq\varnothing$, we get $g(\Omega)=D$, i.e. $g$ is surjective.

Now $g:\Omega\to D$ is a surjective proper local homeomorphism, hence a covering map. Because $D$ is a Jordan domain, it is simply connected. Any connected covering over $D$ has one sheet. As such, $g$ is a homeomorphism $\Omega\to D$.

Finally, we already have:
\[
f|_{\Omega}:\Omega\to D \text{ is a homeomorphism},\qquad
f|_{\partial\Omega}:\partial\Omega\to\Gamma \text{ is a homeomorphism},
\]
and $D\cap\Gamma=\varnothing$. Hence $f:\overline\Omega\to\overline D$ is bijective. It is continuous by assumption, and $\overline\Omega$ is compact while $\overline D$ is Hausdorff. A continuous bijection is a homeomorphism. Therefore,
$f:\overline{\Omega}\to\overline D\text{ is a homeomorphism}.$
\end{proof}
\section{Circular Quasi-Conformal Deturbulence}

\subsection{Circular Turbulence Removal}

A crucial challenge in learning-based turbulence removal is the absence of turbulence-free label data for supervised training. This is evident, as a distorted image and its distortion-free counterpart cannot coexist at the same time and in the same scene. Therefore, the primary objective of this work is to develop an unsupervised model for turbulence removal. To achieve this, we design our model with a circular architecture, allowing it to evolve iteratively in a circular manner.

\begin{figure*}
    \centering
    \includegraphics[width=1\linewidth]{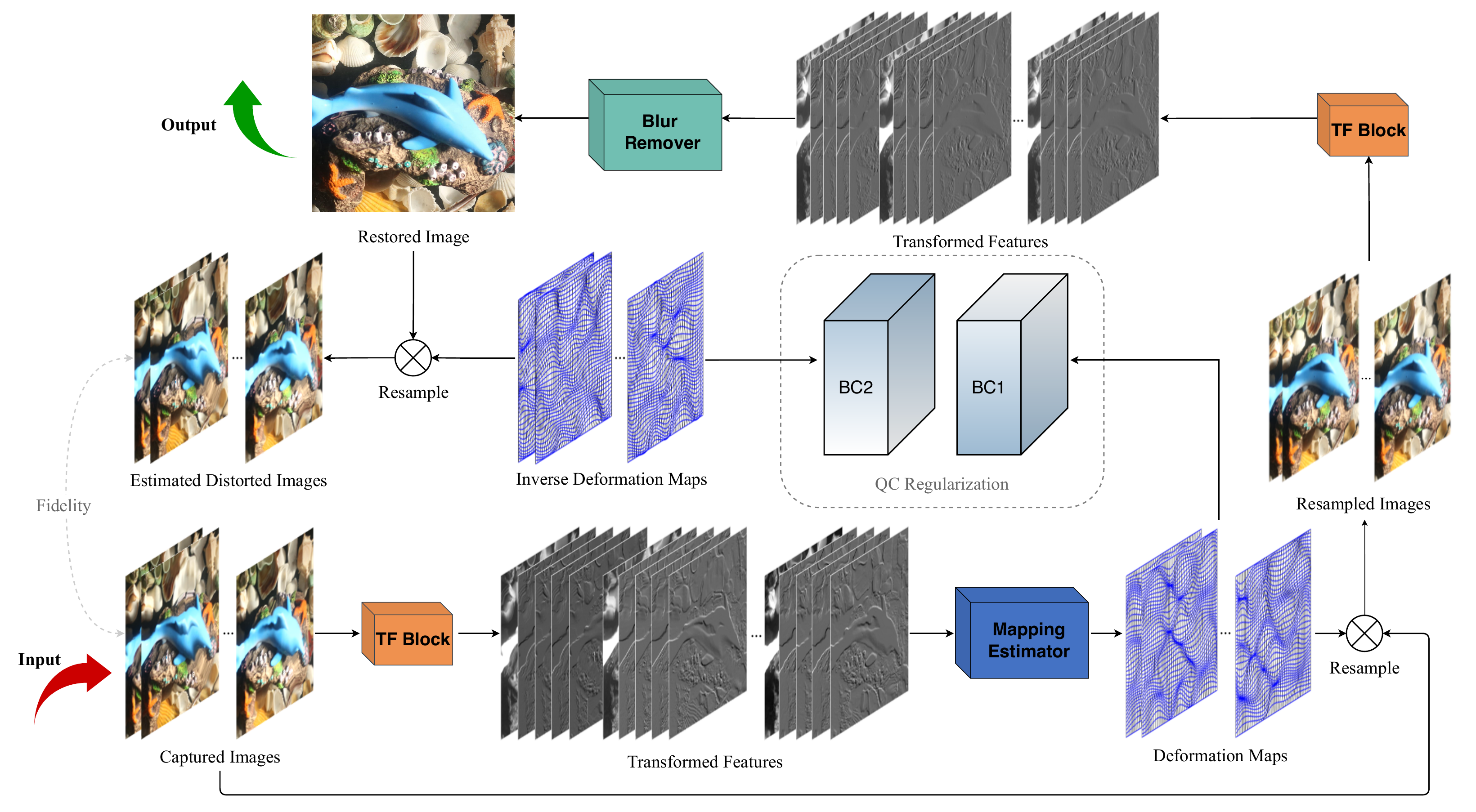}
    \caption{Illustration of the proposed Circular Quasi-Conformal Deturbulence (CQCD) framework. First, a deformation mapping is estimated to geometrically correct the captured distorted image using the Mapping Estimator, which takes the captured image and its tight-frame (TF) features extracted by the TF block. A pixel-wise network then removes both existing blurs and those introduced during resampling via the Blur Remover, which processes the resampled image along with its TF features. The restored image is then mapped back using the inverse deformation to reconstruct a distorted image, which is compared to the original via a fidelity loss, completing the cycle. The QC regularization modules compute the Beltrami coefficients for both the forward and inverse deformation maps.}
    \label{fig:framework}
\end{figure*}

Here, we describe the entire algorithm in three stages: distortion-free image estimation, restored image estimation, and distorted image estimation. In the distortion-free image estimation stage, a mapping is estimated to geometrically restore the captured distorted image. Since the distortion-free image is obtained through resampling via interpolation using the estimated mapping, the resampled image inevitably contains blurs introduced by the interpolation process. To address this, a pixel-to-pixel network is employed to remove these blurs, producing a deblurred image, which serves as the desired restored image. In the final stage, distorted image estimation, the inverse of the previously estimated mapping is applied to the restored image to reproduce distorted images similar to the originally captured ones. This ensures consistency and completes the cycle. The details of the entire algorithm are presented step by step below.

In the first stage, a sequence of captured distorted images, ${\tilde{I}_t}$, is provided as input. To process these images, we first apply a tight-frame block (Section~\ref{tf_trans}) to extract transformed features from the distorted image sequence. This transformation effectively captures essential structural information while mitigating noise and artifacts present in the original inputs. The transformed features are computed as:
\begin{equation}
\tilde{F}_t = \mathcal{F}( \tilde{I}_t )
\end{equation}
where $\mathcal{F}$ represents the tight-frame transformation. These extracted features serve as input to the subsequent module, which is responsible for estimating the geometric deformation maps necessary for restoring the original undistorted image.

A neural network, referred to as the deformation estimator $\mathcal{N}_{DE}$, is utilized to predict a mapping that rectifies the geometric distortions present in the input images. The deformation map for each image frame is computed as:
\begin{equation}
f_t = \mathcal{N}_{DE} (\tilde{F}_t)
\end{equation}
where $f_i$ denotes the estimated geometric distortion field for the $i$-th frame. It takes the tight-frame features $\tilde{F}_t$ as input and outputs a deformation map $f_t$ represented as a two-channel displacement field, consisting of horizontal and vertical displacement values for each position. Note that the mappings are designed to be topologically consistent and practically invertible for the later stages of the proposed circular model, where inverse mappings are required. This property is encouraged through a regularization term based on the Beltrami coefficient, which will be detailed in a later section.

Using the estimated deformation maps, we perform resampling to warp the distorted images accordingly:
\begin{equation}
\hat{I}_t = \tilde{I}_t \circ f_t
\end{equation}

The resampled images $\hat{I}_t$ ideally approximate a geometrically undistorted version of the original frames. At this point, we complete the first stage, obtaining the resampled images in which geometric turbulence has been mitigated. However, due to the resampling process, additional blurring artifacts may be introduced, together with the blur in the original images, necessitating further refinement in the subsequent stages.
Next, we proceed to the second stage. To further refine the image representation, we apply another tight-frame transformation to the resampled images:
\begin{equation}
\hat{F}_t = \mathcal{F}( \hat{I}_t )
\end{equation}

These transformed features serve as the input for a network, namely blur remover, denoted as $\mathcal{N}_{BR}$, which synthesizes the final restored image:
\begin{equation}
I^* = \mathcal{N}_{BR} (\hat{F}_1,\dots,\hat{F}_T)
\end{equation}

The blur remover is trained to ensure that the output $I$ is not only geometrically corrected but also visually sharp. Blurring in the resampled image may arise from both the original captured frames and interpolation during the resampling process. By leveraging the blur remover module and the complementary information across multiple frames, these residual blur effects can be effectively mitigated. This is achieved by minimizing the following reconstruction loss:
\begin{equation}
\mathcal{L}_{rec} = \frac{1}{T}\sum_{t=1}^{T}|| I^* - \hat{I}_t ||_1
\end{equation}

At this point, the second stage is completed, and the resulting image $I$ is the desired restored image.

To validate the correctness of the estimated restored image, we check whether the geometric distortions in the input frames can be accurately reproduced using the restored image and the estimated distortion mappings. {The idea is that, since the mapping $f_t$ is used to geometrically correct the distorted images, its inverse naturally represents the distortion process.} Therefore, the restored image $I^*$ should be able to reconstruct the original distorted images when the inverse of the estimated deformation maps{, denoted by $f_t^{-1}$,} is applied:
\begin{equation}
\bar{I}_t = I^* \circ f_t^{-1}
\end{equation}
Here, $\bar{I}_t$ then represents the estimated distorted images, which are expected to be structurally similar to the captured distorted images $\tilde{I}_t$. By enforcing this constraint, we ensure that the estimated deformation maps not only remove distortions but also allow for a reversible restoration process. In particular, if the network correctly estimates a deformation mapping $f_t$ that eliminates distortions, applying its inverse $f_t^{-1}$ to the restored image should reconstruct the original distorted frames. This cyclic property enhances structural consistency and ensures that the estimated deformations faithfully correspond to the actual distortions present in the input images.

Additionally, since the mappings are expected to be invertible in the circular loop, the network is guided to learn smooth, approximately homeomorphic deformations rather than abrupt distortions. Enforcing this constraint helps the restored images retain their original structural integrity, preventing excessive warping and the introduction of geometric artifacts. This not only enhances the realism of the restored images but also improves consistency in the overall transformation process.

For consistency, we connect the third stage with the first stage to complete the cycle. This is achieved by computing a similarity fidelity loss between the estimated distorted images $\bar{I}_t$ and the original captured images $\tilde{I}_t$ as follows:
\begin{equation}
\mathcal{L}_{dist} = \frac{1}{T}\sum_{t=1}^{T}|| \tilde{I}_t - \bar{I}_t ||_1
\end{equation}
This loss promotes the accuracy of the estimated deformation fields and helps the restoration process maintain structural integrity throughout the cycle. By requiring that the estimated distorted images, $\bar{I}_t = I^* \circ f_t^{-1}$, closely match the original distorted images $\tilde{I}_t$, the model benefits from an implicit supervision signal. This self-supervised approach enables the network to learn meaningful deformations without the need for paired training data, which is often unavailable in real-world scenarios.

{
Minimizing $\mathcal{L}_{rec}$ alone may lead to a trivial solution in which the restored image $I^*$ becomes the pixel-wise mean of the resampled frames $\{\hat{I}_t\}_{t=1}^T$, resulting in blur and ghosting artifacts caused by frame misalignment. The distortion loss $\mathcal{L}_{dist}$ mitigates this issue by requiring the re-distorted reconstructions $\bar{I}_t = I^* \circ f_t^{-1}$ to reproduce the frame-specific distortions observed in $\tilde{I}_t$. Consequently, the blur remover is encouraged to produce a cleaner and sharper estimate $I^*$ that faithfully represents the underlying scene rather than collapsing to a trivial average.
}

\subsection{Tight-Frame Encoding}
\label{tf_trans}

\begin{figure*}
    \centering
    \includegraphics[width=0.7\textwidth]{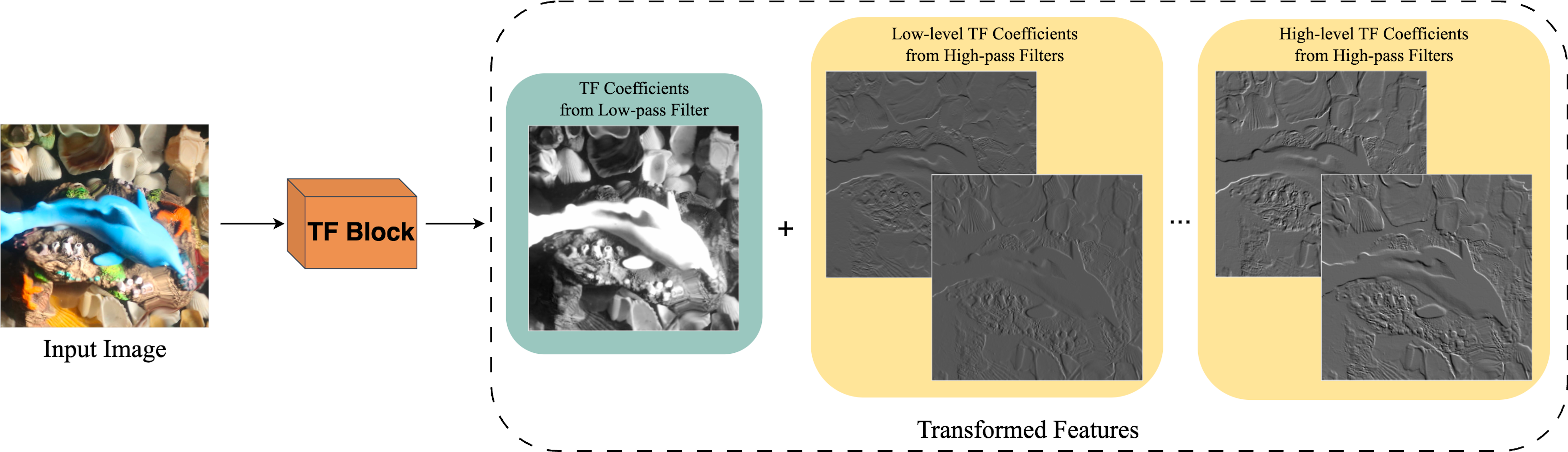}
    \caption{Schematic illustration of the Tight-Frame (TF) Block.}
    \label{fig:tfblock}
\end{figure*}

In this subsection, we will briefly introduce the concept of tight frames and their integration, referred to as TF (tight-frame) Block, shown in Figure~\ref{fig:tfblock}, into our framework. The TF Block is designed to explicitly separate horizontal and vertical frequency components through edge variations, enabling more effective capture of common geometric structures across feature channels~\cite{chen2024mocha}. More detailed theories and applications regarding tight frames can be found in~\cite{ron1997affine,daubechies2003framelets,chan2004tight,shen2010wavelet}.

The tight frame transformation is constructed from a one-dimensional low-pass filter $m_0$ and some one-dimensional high-pass filters $\{m_p\}_{p=1}^r$. Those filters are constructed under the Unitary Extension Principle (UEP)~\cite{ron1997affine}:
\begin{equation}
    \sum_{p=0}^r|\hat{m}_p(\xi)|^2 = 1, \quad \forall \xi \in \mathbb{R}.
    \label{eq:UEP}
\end{equation}
where $\hat{m}_p$ denotes the Fourier transform of $m_p$. 

In our implementation, we set \( r = 2 \) and adopt the following filters: a low-pass filter \( m_0 = \frac{1}{3}[1, 1, 1] \) to capture low-frequency components, and two high-pass filters \( m_1 = \frac{\sqrt{6}}{6}[1, 0, -1] \) and \( m_2 = \frac{3\sqrt{2}}{18}[1, -2, 1] \) to extract multi-directional high-frequency details. Notably, \( m_1 \) corresponds to a scaled first-order central difference operator, while \( m_2 \) approximates a scaled second-order central difference, both in the one-dimensional setting.

For the 2D case, the forward and inverse tight frame filters \( \{\mathcal{W}_{p,q}\}_{p,q=0}^r \) and \( \{\tilde{\mathcal{W}}_{p,q}\}_{p,q=0}^r \) are constructed using tensor products:
\begin{equation}
\mathcal{W}_{p,q} = m_p \otimes m_q^{\intercal}, \quad \tilde{\mathcal{W}}_{p,q} = \tilde{m}_p \otimes \tilde{m}_q^{\intercal},
\end{equation}
where \( \tilde{m}_p \) denotes the adjoint filter of \( m_p \), as described in~\cite[Section 2.3]{chan2005resolution}, and \( \otimes \) is the Kronecker product.

Given an image \( I \), the low-frequency subband is obtained by
\begin{equation}
\mathcal{W}_{0,0}(I) = \mathcal{W}_{0,0} * I,
\end{equation}
while the high-frequency directional subbands are obtained as
\begin{equation}
\mathcal{W}_{p,q}(I) = \mathcal{W}_{p,q} * I \quad \text{for } (p,q) \neq (0,0),
\end{equation}
where \( * \) denotes convolution. These subbands capture directional textures and edges essential for detailed image analysis and restoration.

Furthermore, the tight frame transform supports a multi-level decomposition mechanism, enabling the extraction of progressively finer and more detailed feature maps. 
We define the level-$L$ tight framelets $F^L$ as:
\begin{equation}
\begin{aligned}
    F^L = \{C^L\} &\cup \{H_{p,q}^l \text{ for $(p,q)\neq(0,0)$, $l=1,\dots,L$}\},
\end{aligned}
\end{equation}
where $C^L$ and $H^L$ denote the low-frequency and high-frequency subband calculated at the level $L$th decomposition, respectively. The subbands can be constructed via multiple convolution operations on image $I$, formulated as:
\begin{equation}
\begin{aligned}
    & C^L = (\mathcal{W}_{0,0})^L (I), \\
    & H_{p,q}^L = \mathcal{W}_{p,q} \left((\mathcal{W}_{0,0})^{L-1}(I)\right) \quad \text{for } \hfill (p,q) \neq (0,0),\\
\end{aligned}
\end{equation}



It is also important to note that the collection of features extracted through the tight-frame transform is sufficient for perfect reconstruction \cite[Section 3.2]{chan2004tight}:
\begin{equation}
\begin{aligned}
    I = &\tilde{\mathcal{W}}_{0,0}(C^L) + \sum_{l=1}^L(\tilde{\mathcal{W}}_{0,0})^{l-1}\left( \sum_{\substack{p,q=0 \\ (p,q)\neq (0,0)}}^{r}\tilde{\mathcal{W}}_{p,q}(H_{p,q}^l)\right), \\
      = &\tilde{\mathcal{W}}_{0,0}\left((\mathcal{W}_{0,0})^L (I)\right) + 
      \sum_{l=1}^L(\tilde{\mathcal{W}}_{0,0})^{l-1}\left( \sum_{\substack{p,q=0 \\ (p,q)\neq (0,0)}}^{r}\tilde{\mathcal{W}}_{p,q} *\mathcal{W}_{p,q} * (\mathcal{W}_{0,0})^{l-1}(I)\right),
\end{aligned}
\end{equation}
where $(\mathcal{W}_{p,q})^l$ and $(\tilde{\mathcal{W}}_{p,q})^l$ denote the order $l$ convolution operations of the forward and its inverse tight frame filters, respectively.

This property implies that the tight-frame coefficients serve as a complete representation of the original image $I$, preserving all essential information without loss of detail. Theoretical guarantees of perfect reconstruction provide a solid foundation for using tight-frame transforms in image restoration tasks.

Moreover, by applying the transform across multiple levels, features are extracted into increasingly explicit and structured forms. These enriched representations are more interpretable for neural networks and offer additional redundancy beyond the original image. The multi-level tight-frame transformation can then enhance robustness against noise and minor geometric deformations.

The tight-frame transform inherently encodes explicit geometric semantics. Its directional sensitivity stems from the construction of filters, which effectively separate horizontal and vertical high-frequency components. To illustrate this more concretely, consider the two example filters below:
\begin{equation}
\begin{aligned}
    \mathcal{W}_{0,1} = \frac{\sqrt{6}}{18}
    \begin{pmatrix}
        1 & 1 & 1\\
        0 & 0 & 0\\
        -1& -1& -1
    \end{pmatrix},
    \mathcal{W}_{1,0} = \frac{\sqrt{6}}{18}
    \begin{pmatrix}
        \;1 & \;\;\;0 & \;-1\\
        \;1 & \;\;\;0 & \;-1\\
        \;1 & \;\;\;0 & \;-1
    \end{pmatrix}.
\end{aligned}
\end{equation}
Here, $\mathcal{W}_{0,1}$ resembles a scaled version of the 2D first-order central difference operator in the vertical direction, while $\mathcal{W}_{1,0}$ corresponds to the horizontal direction. The scaling ensures these filters satisfy the UEP in Equation \eqref{eq:UEP}.

Other filters $\mathcal{W}_{p,q}$ follow a similar design, capturing features in different directions. As a result, tight-frame transforms can extract multi-directional features, making structured patterns such as edges and straight lines more explicit in the input representation fed to the neural network.

In this sense, horizontal edge misalignments in turbulent images are predominantly captured in $\{H_{p,0}^l\}_{p=1}^{2}$, and vertical jittering appears in $\{H_{0,q}^l\}_{q=1}^{2}$, while $H_{1,1}^l$ and $H_{2,2}^l$ show diagonal deformation and Laplacian, respectively.

Assuming the deformation field to be $\triangle = (\triangle_x, \triangle_y)^{\intercal}$, and by Taylor approximation of the turbulence-corrupted image formula in Equation \eqref{eq:distort_formula}:
\begin{equation}
    \tilde{I}(x,y) = I(x,y) + \nabla I(x,y) \cdot \triangle + \mathcal{O}(||\triangle||^2),
\end{equation}
and apply tight frame transform for both sides,
\begin{equation}
\begin{aligned}
    \tilde{H}_{p,q}^l \approx H_{p,q}^l + \nabla {H_{p,q}^l} \cdot \triangle,
\end{aligned}
\end{equation}
where $\tilde{H}_{p,q}^l$ represents high-frequency subbands decomposed from distorted image $\tilde{I}$.

With the directional anisotropic tight frame transform (e.g., horizontal and vertical), we also have:
\begin{equation}
    \frac{\partial H_{p,0}^l}{\partial \triangle_x} \gg \frac{\partial H_{p,0}^l}{\partial \triangle_y}, \quad \frac{\partial H_{0,q}^l}{\partial \triangle_y} \gg \frac{\partial H_{0,q}^l}{\partial \triangle_x}, \quad p,q=1,2.
\end{equation}
Similar properties can be drawn analogously for other directions. Thus, the tight frame features exhibit high directional sensitivity, as can be observed from Figure~\ref{fig:tfblock}, allowing CNNs to directly access orientational deformation patterns without learning fundamental frequency representations compared to Fourier-based encoding, thereby improving deformation field estimation accuracy. The multi-level transform further disentangles image gradients across scales, helping CNNs hierarchically resolve deformation magnitudes—fine-grained distortions at shallow layers and coarse structural shifts at deeper layers.

\subsection{Homeomorphic Mapping Estimation}

\begin{figure*}
    \centering
    \includegraphics[width=0.8\textwidth]{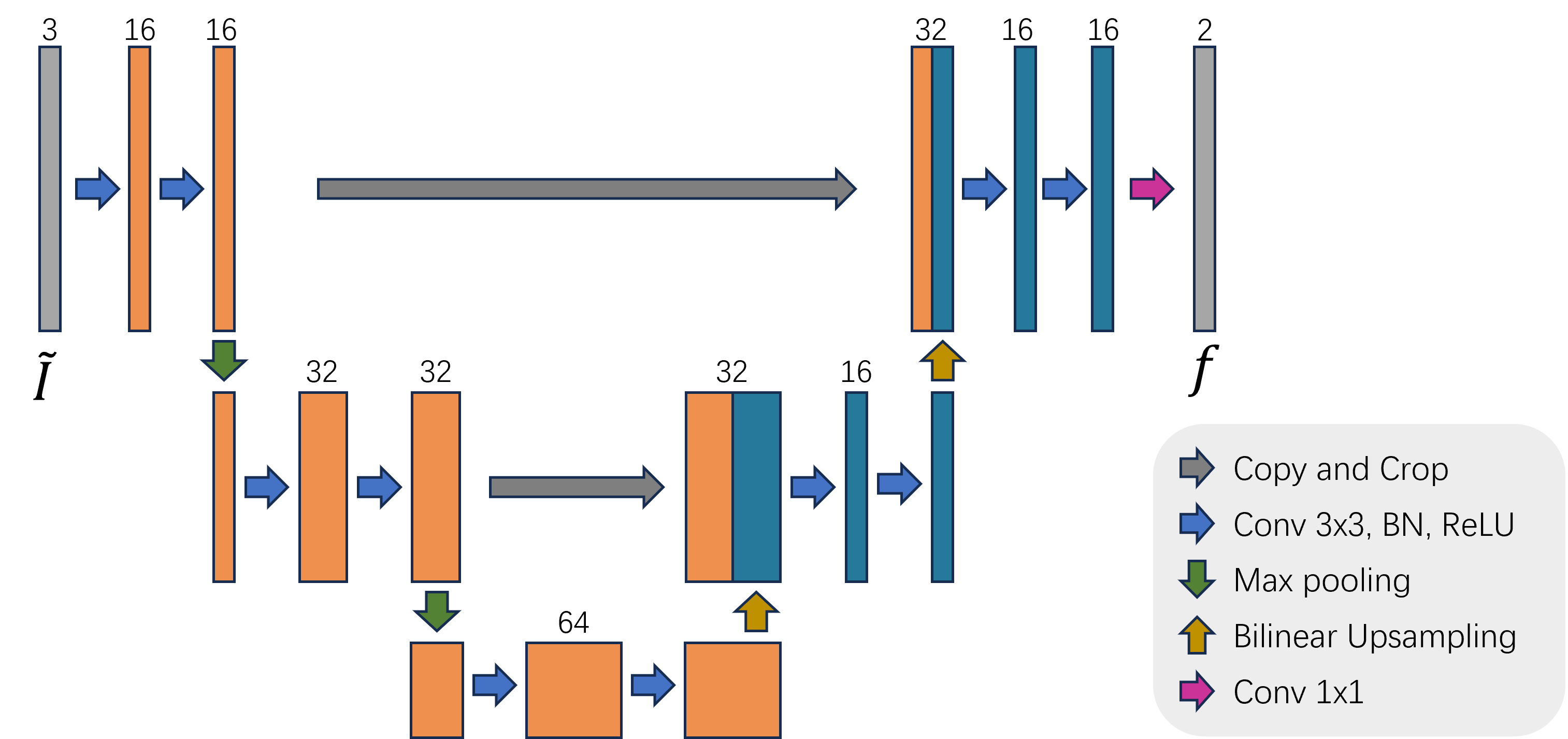}
    \caption{Illustration of Deformation Estimator.}
    \label{fig:deformestimator}
\end{figure*}

Homeomorphic mapping is crucial for obtaining smooth and physically meaningful warping without introducing folds. If the estimated deformation field contains folds, the mapping becomes non-injective, leading to structural inconsistencies and information loss, ultimately degrading the restoration quality. Therefore, to faithfully recover the original image structure and avoid excessive warping or physically implausible artifacts, the deformation mapping is encouraged to be homeomorphic.

Moreover, the proposed circular architecture relies on both the estimated deformation mapping $f_i$ to correct distortions and its inverse $f_i^{-1}$ to reconstruct the distorted inputs. This bidirectional process requires stable practical invertibility. Loss of invertibility directly contradicts the core principle of the circular design, which aims to progressively refine the restoration by enforcing mutual consistency. As such, the homeomorphism property should be promoted and explicitly reinforced to support the integrity and effectiveness of the circular framework.

To encourage homeomorphic behavior, the deformation estimator, implemented as a UNet (Figure~\ref{fig:deformestimator}), is regularized to produce deformation fields with small Beltrami coefficients. Theorem~\ref{thm:qcbc} provides the continuous-domain motivation for this design under bounded-domain assumptions. The network outputs a two-channel displacement field, where each channel encodes the horizontal and vertical displacement at every pixel. Together, these two channels provide a complete and differentiable parametrization of the spatial transformation, as illustrated in Figure~\ref{fig:image_map_composition}. This formulation enables differentiable warping and seamless integration with the subsequent components of the circular pipeline.

To promote this condition and smoothness of the mapping, we introduce the following additional QC regularization terms:
\begin{equation}
\label{eq:bcloss}
\begin{aligned}
\mathcal{L}_{\text{bc}} &= \frac{1}{4T}\sum_{t=1}^T\left(||\mu(f_t)||_2^2 + ||\mu(f_t^{-1})||_2^2 \right),
\end{aligned} 
\end{equation} 
where $\mu(f)$ represents the Beltrami coefficient of the mapping $f$, computed using a Finite Difference Method implementation of Equation \eqref{eq:beleq}. This QC regularization is inspired by Theorem~\ref{thm:qcbc}: by minimizing a properly weighted $\mathcal{L}_{\text{bc}}$, we encourage small Beltrami coefficients and empirically bijective deformations in the learned mappings. In this way, the forward mapping corrects geometric distortions, while the inverse mapping reconstructs the original distortions for cycle-consistency verification. Symmetrically constraining both mappings in Eq.~(\ref{eq:bcloss}) promotes quasi-conformal consistency throughout the cycle, empirically reduces foldings, and stabilizes optimization. Empirical results show that this regularization improves invertibility behavior in practice.


Overall, the framework of CQCD is shown in Figure~\ref{fig:framework}, where the deformation estimator $\mathcal{N}_{DE}$ and the blur remover $\mathcal{N}_{BR}$ are optimized by minimizing $\mathcal{L}_{DE}$ and $\mathcal{L}_{BR}$, respectively, which are designed as
\begin{equation}
\mathcal{L}_{DE} = \mathcal{L}_{dist} + \mathcal{L}_{rec} + \lambda \mathcal{L}_{\text{bc}},\quad \mathcal{L}_{BR} = \mathcal{L}_{dist} + \mathcal{L}_{rec},
\end{equation}
where the BC weighting $\lambda$ controls the trade-off between the fidelities and regularization.
\section{Experiments}
In this section, we evaluate the effectiveness of the proposed CQCD framework through a series of experiments. Specifically, we conduct tests on synthetic turbulence videos with varying frame counts, as well as real turbulence-distorted images, comparing our results with those of state-of-the-art (SOTA) methods. Additionally, we compare the deformation fields estimated by our model against the ground-truth fields used to simulate turbulence, further validating the accuracy of our approach. Finally, we perform ablation studies on the weighting of the quasi-conformal (QC) regularization and the impact of incorporating the tight-frame (TF) block.

\subsection{Implementation Details}
We begin by detailing the experimental settings, such as the network settings and data sources used in this work. and the computational environment.

\textbf{Architecture Detail} 
The architecture of the deformation estimator is a U-net-shaped CNN displayed in Figure~\ref{fig:deformestimator}. The blur remover is structured as a sequential stack of 1×1 convolutional layers with two hidden layers. The architecture begins with an initial 1×1 convolutional layer that maps the input TF features to 256 channels, followed by a ReLU activation and batch normalization, while the network ends with sigmoid non-linearity for the restored image. Two networks are optimized alternatively every 100 epochs with the $\mathcal{N}_{BR}$ trained for 50 epochs using the Adam optimizer and the $\mathcal{N}_{DE}$ trained for another 50 epochs using the RMSprop optimizer. {The inverse mapping is computed using the function \texttt{invert\_flow}~\cite{ravasio_oflib}, which supports gradient propagation through the circular pipeline.}


\textbf{Data Setup}
\begin{figure*}
    \centering
    \includegraphics[width=0.75\textwidth]{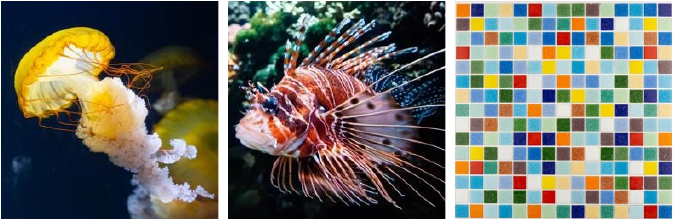}
    \caption{Reference images of underwater scenes, \textit{Jellyfish} (mild), \textit{Pterois} (medium) and \textit{Tiles} (severe), used for simulations.}
    \label{fig:GTImages}
\end{figure*}
Our experiments utilize both synthetic turbulence-distorted images and real-world images corrupted by turbulence to validate the proposed framework. For synthetic data, we generate distorted images from publicly available underwater photographs by applying the turbulence simulation method from Thapa et al.~\cite{thapa2020dynamic}. While retaining the default parameters of the dynamic turbulence model, we adjust two key variables to simulate varying turbulence intensities: we set $\alpha=12$ and level factor to 18 for mild turbulence, while $\alpha=13$ and level factor to 20 for the median case, and severe turbulence is simulated with $\alpha=15$ and level factor of 25. These parameters are applied to three representative underwater scenes (\textit{Jellyfish}, \textit{Pterois}, and \textit{Tiles}), as in Figure~\ref{fig:GTImages}. For real-world evaluation, we adopt turbulence-corrupted images from the publicly available datasets of Li et al.~\cite{li2018learning}, Tian et al.~\cite{tian2009seeing}, Alterman et al.~\cite{alterman2012detecting}, and Mao et al.~\cite{mao2020image}, focusing on their demo subsets that capture diverse turbulence patterns in aquatic environments. 

\textbf{Computational Resources}
All experiments were conducted on a PC equipped with Intel\textsuperscript{\textcircled{\textsc{r}}} Xeon\textsuperscript{\textcircled{\textsc{r}}} Silver 4210 Processor CPU 2.20GHz and Nvidia GeForce RTX 3090 GPU with 24G of memory. Under our default configuration, the GPU memory consumption during training is approximately 7 GB for a 10-frame sequence with 256$\times$256 of spatial resolution. The total optimization time for 2000 iterations in this setting is $\sim$206 seconds, as referenced in the ablation studies (Table~\ref{tab:ablation_studies}), which further illustrates how this cost varies with architectural choices.

\subsection{Comparison with Unsupervised Methods Across Different Frame Counts}

To validate the effectiveness of our CQCD framework, we first compare it against a simple average approach and two SOTA unsupervised water turbulence removal methods: NeRT~\cite{li2021unsupervised} and NDIR~\cite{jiang2023nert}. The evaluation is conducted across three turbulence levels using the simulated underwater images. For each level, we vary the number of input distorted frames (5, 10, 15) to assess the robustness of different numbers of input frames.

\begin{table*}[ht]
\small
    \centering
    \begin{tabular}{ c c c c c c c c c c}
        \hline
        Image & \multicolumn{3}{c}{Jellyfish (Mild)} & \multicolumn{3}{c}{Pterois (Medium)} & \multicolumn{3}{c}{Tiles (Severe)} \\
        \#Input & 5 & 10 & 15 &  5 & 10 & 15 & 5 & 10 & 15 \\  
        \hline
        \multirow{2}{*}{Average} & 23.24 & 24.79 & 25.45 & 17.02 & 17.32 & 17.54 & 15.21 & 15.47 & 15.66 \\
        & 0.8480 & 0.8721 & 0.8926 & 0.6499 & 0.6631 & 0.6687 & 0.6169 & 0.6290 & 0.6357 \\
        \multirow{2}{*}{TurbRecon} & 15.41 & 15.42 & 15.42 & 13.27 & 13.28 & 13.29 & 13.77 & 13.78 & 13.80 \\
         & 0.6026 & 0.6462 & 0.6843 & 0.5411 & 0.5673 & 0.6031 & 0.5420 & 0.5544 & 0.5702 \\
         \multirow{2}{*}{DiffTemplate} & 23.53 & 27.01 & 27.72 & 18.37 & 18.50 & 18.55 & 16.48 & 16.93 & 17.36 \\
         & 0.8420 & 0.8943 & 0.9092 & 0.6727 & 0.6938 & 0.6903 & 0.6412 & 0.6695 & 0.6933 \\
        \multirow{2}{*}{NeRT} & 20.63 & 23.51 & 24.47 & 14.77 & 15.53 & 17.22 & 14.86 & 15.18 & 17.98 \\
         & 0.8492 & 0.8991 & 0.9120 & 0.6106 & 0.6464 & 0.7197 & 0.7031 & 0.7410 & 0.7708 \\
         \multirow{2}{*}{NDIR} & 22.47 & 26.15 & 28.00 & 17.55 & 17.98 & 20.55 & 18.14 & 19.61 & 21.79 \\
         & 0.8306 & 0.8909 & 0.9118 & 0.7573 & 0.7837 & 0.8615 & 0.7566 & 0.8126 & 0.8680 \\
        \multirow{2}{*}{Ours} & \textbf{23.70} & \textbf{27.02} & \textbf{28.43} & \textbf{18.64} & \textbf{18.74} & \textbf{20.93} & \textbf{19.07} & \textbf{20.16} & \textbf{22.40} \\
         & \textbf{0.8506} & \textbf{0.9013} & \textbf{0.9127} & \textbf{0.7921} & \textbf{0.8039} & \textbf{0.8668} & \textbf{0.7912} & \textbf{0.8212} & \textbf{0.8765} \\
        \hline
    \end{tabular}
\caption{Quantitative evaluation on water turbulence. Each double-row entry reports PSNR$\uparrow$ (top, in dB) and SSIM$\uparrow$ (bottom). Bold values indicate the best result for each sample.}
\label{synimg_result}
\end{table*}

\begin{figure*}
    \centering
    \includegraphics[width=1\textwidth]{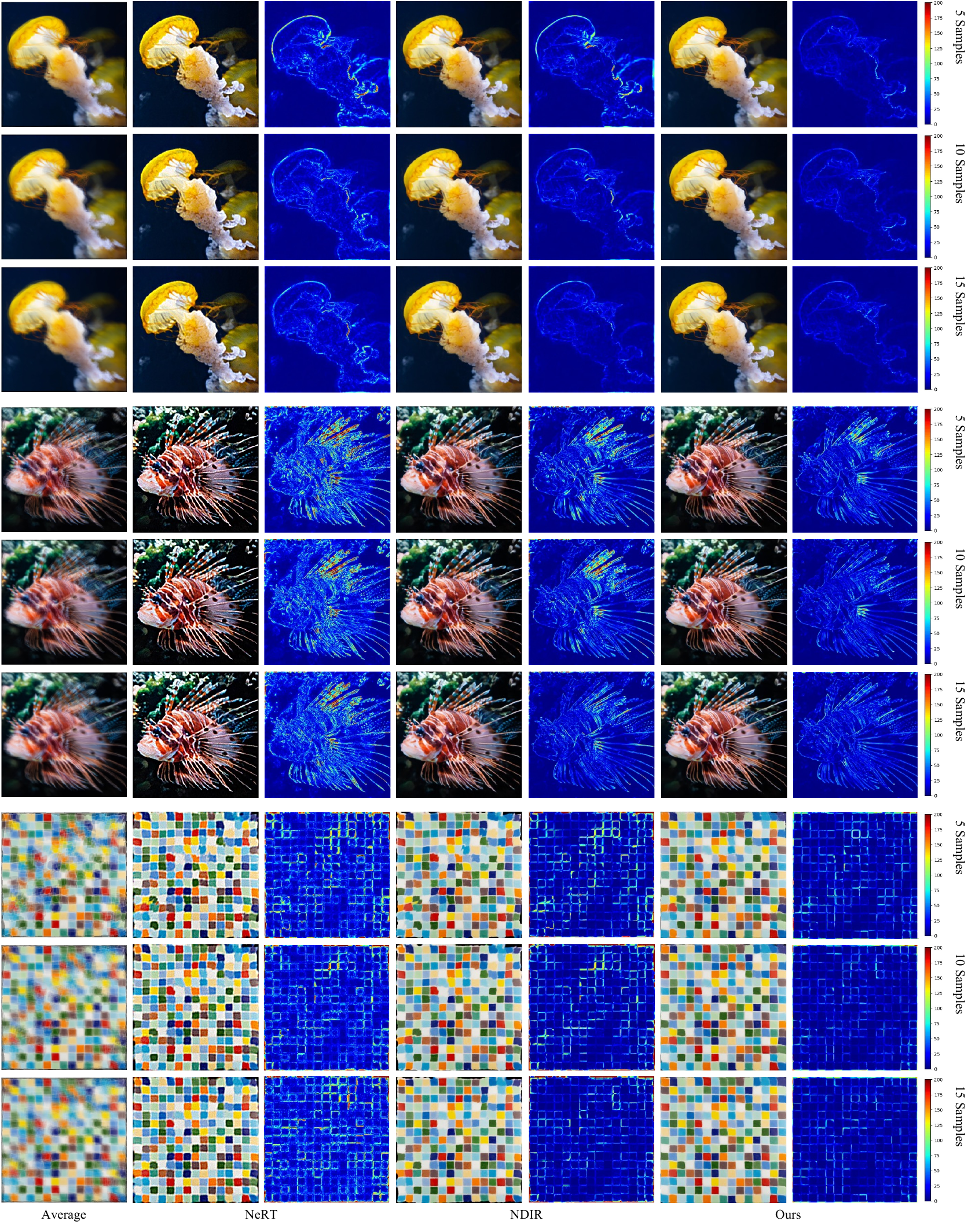}
    \caption{Comparison of unsupervised deturbulence based on \textit{Jellyfish} (mild), \textit{Pterois} (medium) and \textit{Tiles} (severe). From top to bottom, restoration results from 5 distorted images, while the middle from 10, and the right from 15, are presented for different methods, with error maps on the side.}
    \label{fig:ResImages}
\end{figure*}

From the visual comparisons in Figure~\ref{fig:ResImages}, the tentacles of the jellyfish are clearly and accurately restored by our CQCD model, with each tentacle distinctly visible and easily distinguishable. In contrast, NDIR struggles to recover the tentacles when provided with insufficient frames, and NeRT, while enhancing overall sharpness, fails to preserve fine tentacle details.

For the Pterois image affected by medium turbulence, our model successfully preserves fine structures, such as the Pterois fins, even with just 5 input frames, demonstrating its strong capability in handling geometric distortions. This is attributed to the regularization of the estimated mappings into quasi-conformal mappings, which are orientation-preserving and help prevent non-physical warping while maintaining structural consistency. In contrast, the Pterois fins appear disrupted or lost in the results of the other two methods when only 5 or 10 are input. Furthermore, the fin that appears connected in NDIR with 5 and 10 input frames becomes disrupted when 15 frames are used, likely due to the inclusion of low-quality samples that degrade overall performance.

The tile images suffer from severe turbulence. Since the original layout follows a regular square pattern, it serves as an ideal example to showcase geometric restoration capability. From the comparison, it is evident that our CQCD method achieves the best restoration, with fewer artifacts as the number of input frames increases. In contrast, the other two methods exhibit noticeable geometric distortions, highlighting the effectiveness of our circular correction architecture. Moreover, color bleeding, where colors from one tile appear in neighboring tiles of different colors, is occasionally observed in the compared methods. This issue arises from mappings that undergo orientation changes, for which those methods fail to explicitly prevent.

Quantitative results, summarized in Table~\ref{synimg_result}, further support our visual observations. CQCD achieves superior performance in terms of both PSNR and SSIM across all turbulence levels and input frame counts. Notably, the performance margin grows larger with increasing turbulence severity, highlighting the scalability and robustness of our model. These results confirm that the circular quasi-conformal framework promotes geometric consistency as well as better fidelity and structural preservation in highly distorted image restoration tasks.



\subsection{Visual Comparison on Real Captured Sequences}

\begin{figure*}
    \centering
    \includegraphics[width=1\linewidth]{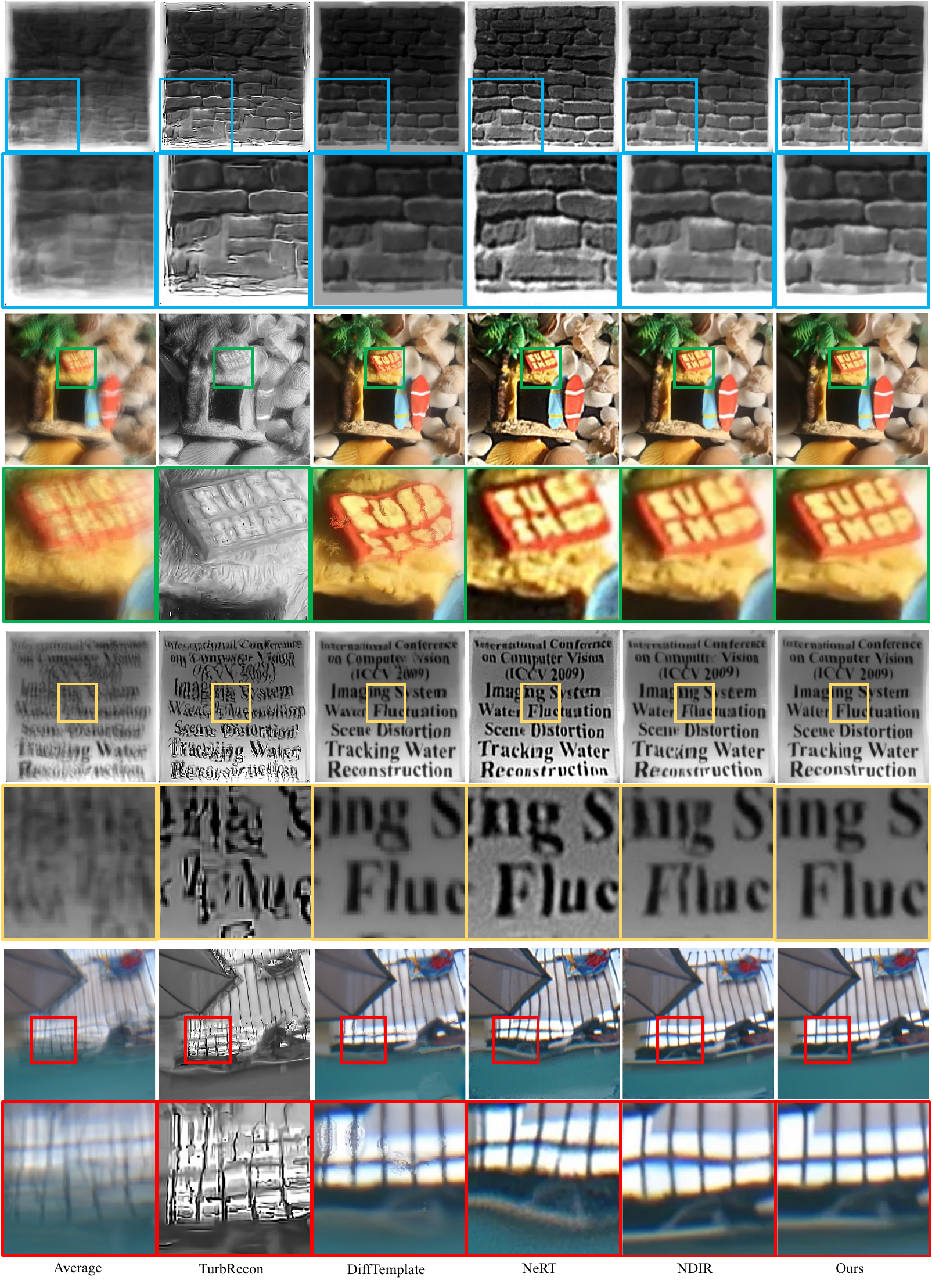}
    \caption{Restoration results of real captured images distorted by water turbulence, with zoomed-in details underneath. Each sample sequence contains 10 images.}
    \label{fig:water}
\end{figure*}

To further validate the generalization capability of CQCD, we benchmark it against two additional deturbulence methods, DiffTemplate~\cite{mao2020image} and TurbRecon~\cite{lao2024diffeomorphic}, alongside the previously evaluated turbulence removal (NeRT, NDIR). All methods are tested on real-world water turbulence-distorted videos, with a fixed input sequence of 10 frames. Since ground truth images are unavailable in real scenarios, we assess performance based on structural plausibility, such as geometric consistency, text readability, and the preservation of straight lines.

As shown in Figure~\ref{fig:water}, the first column shows the average of 10 input frames, which reveals the severity of the water-induced turbulence, characterized by strong geometric distortions, motion blur, and structural deformation. Our CQCD framework restores the brick pattern to a physically correct quasi-rectangular shape, while all other methods leave noticeable bending or warping. In particular, TurbRecon introduces visible artifacts or edge distortions, and DiffTemplate, NeRT, and NDIR fail to completely correct the brick geometry, leaving curves or wavy lines in the result. The ``SURF SHOP'' text on the miniature house is clearly legible in the CQCD result, showing high sharpness and structural coherence. In contrast, DiffTemplate and NDIR produce blurry outputs, with significantly reduced legibility. While NeRT improves sharpness, the text becomes merged and unreadable, likely due to the lack of orientation-aware regularization in the deformation modeling. The third case demonstrates the strongest contrast. CQCD successfully reconstructs each line of text with clean boundaries and spacing. In comparison, TurbRecon suffers from ghosting effects, DiffTemplate and NDIR exhibit overlapping characters, and NeRT shows character bleeding and heavy blur, causing words to become merged or unreadable. This confirms the superior capability of CQCD in restoring fine-grained structures under severe geometric distortion. The image depicts poolside umbrellas and railing structures captured from underwater. CQCD accurately recovers the straight geometry of the railings and the sharp boundaries of the umbrella spokes. Competing methods fail to correct the bending of the rails and introduce inconsistent warping, further emphasizing the strength of CQCD on its circular quasi-conformal design in real-world deturbulence.

\begin{figure*}
    \centering
    \includegraphics[width=1\linewidth]{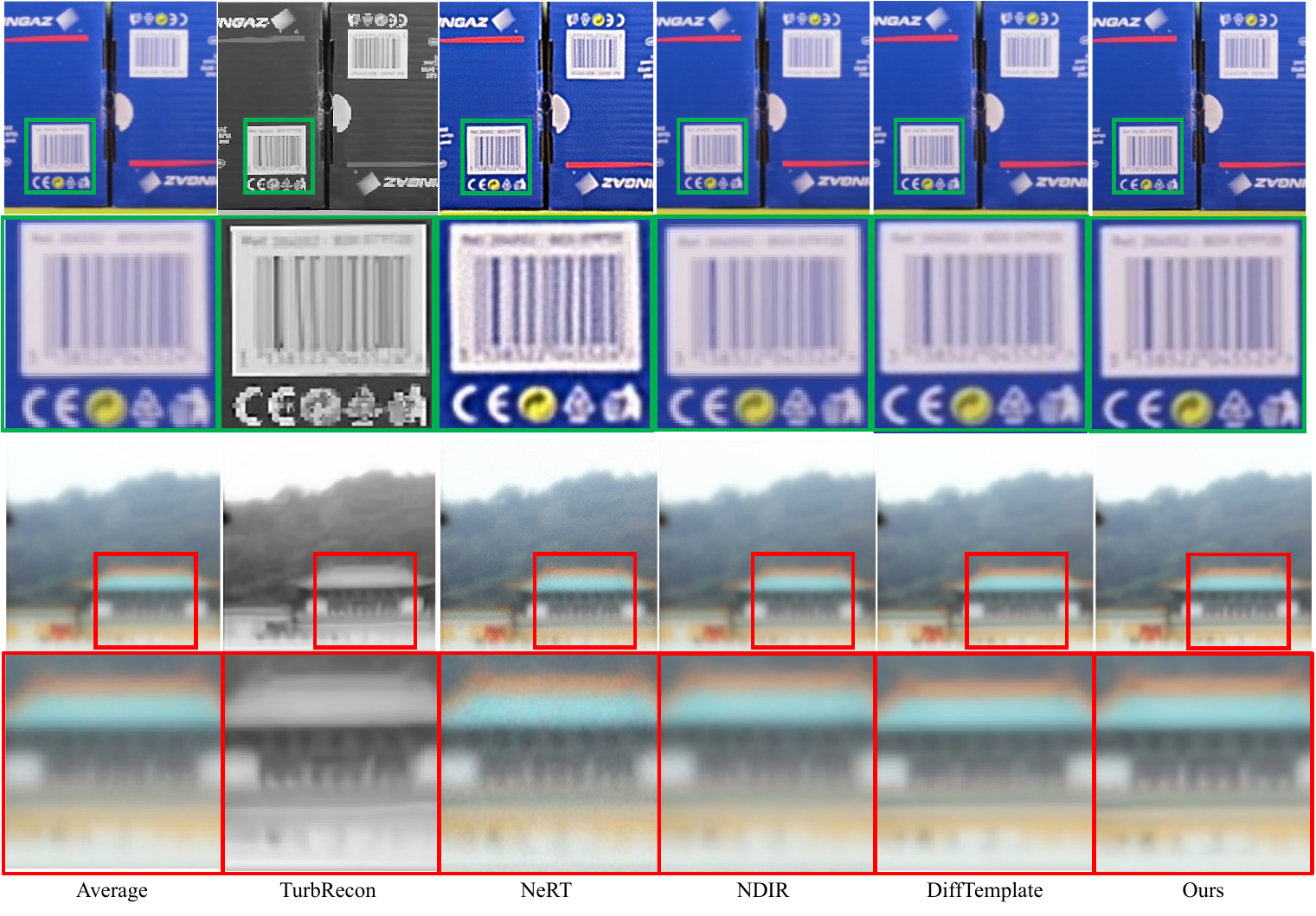}
    \caption{Restoration results of images distorted by air turbulence. Enlarged patches (second/fourth rows) demonstrate the superior capability of our method in recovering clean edges, rectifying geometric distortions, and maintaining structural integrity compared to existing approaches.}
    \label{fig:air}
\end{figure*}

To further assess the capability of CQCD beyond the water turbulence scenarios primarily studied, we evaluate its performance on images distorted by atmospheric turbulence~\cite{mao2020image}, which exhibits distinct spatial and temporal characteristics. Figure~\ref{fig:air} presents restoration results on air turbulence-corrupted imagery, comparing CQCD against several SOTA methods. The visual comparison reveals that CQCD effectively restores geometrically coherent structures and sharp textural details under this different distortion regime. For instance, barcodes, figures, pillars, and railings heavily warped by heat haze are notably rectified, while competing methods often leave residual curvature or artifacts. This demonstrates the geometry-aware correction and self-supervised cycle consistency of CQCD are effective in translating across different turbulence sources.

In summary, CQCD consistently outperforms existing methods in terms of geometric fidelity, clarity, and visual realism. Its ability to suppress distortion while preserving fine structures makes it well-suited for practical underwater and water-surface imaging applications.



\subsection{Restoration and Distortion Field Estimation}

To further validate the rationality and interpretability of the proposed CQCD framework, we evaluate whether the estimated deformation fields produced by the deformation estimator $\mathcal{N}_{DE}$ align with the ground-truth fields used to generate water turbulence-distorted images synthetically. This assessment is critical to understanding whether CQCD accurately captures the underlying geometric distortions induced by turbulence.

We simulated three sets of water turbulence-distorted images, each featuring distinct turbulence patterns by varying flow directions, densities, and spatial distributions. The corresponding ground-truth deformation fields are visualized as distortion grids and exhibit diverse deformation characteristics such as directional shearing, local stretching, and complex warping.

\begin{figure*}
    \centering
    \includegraphics[width=0.83\linewidth]{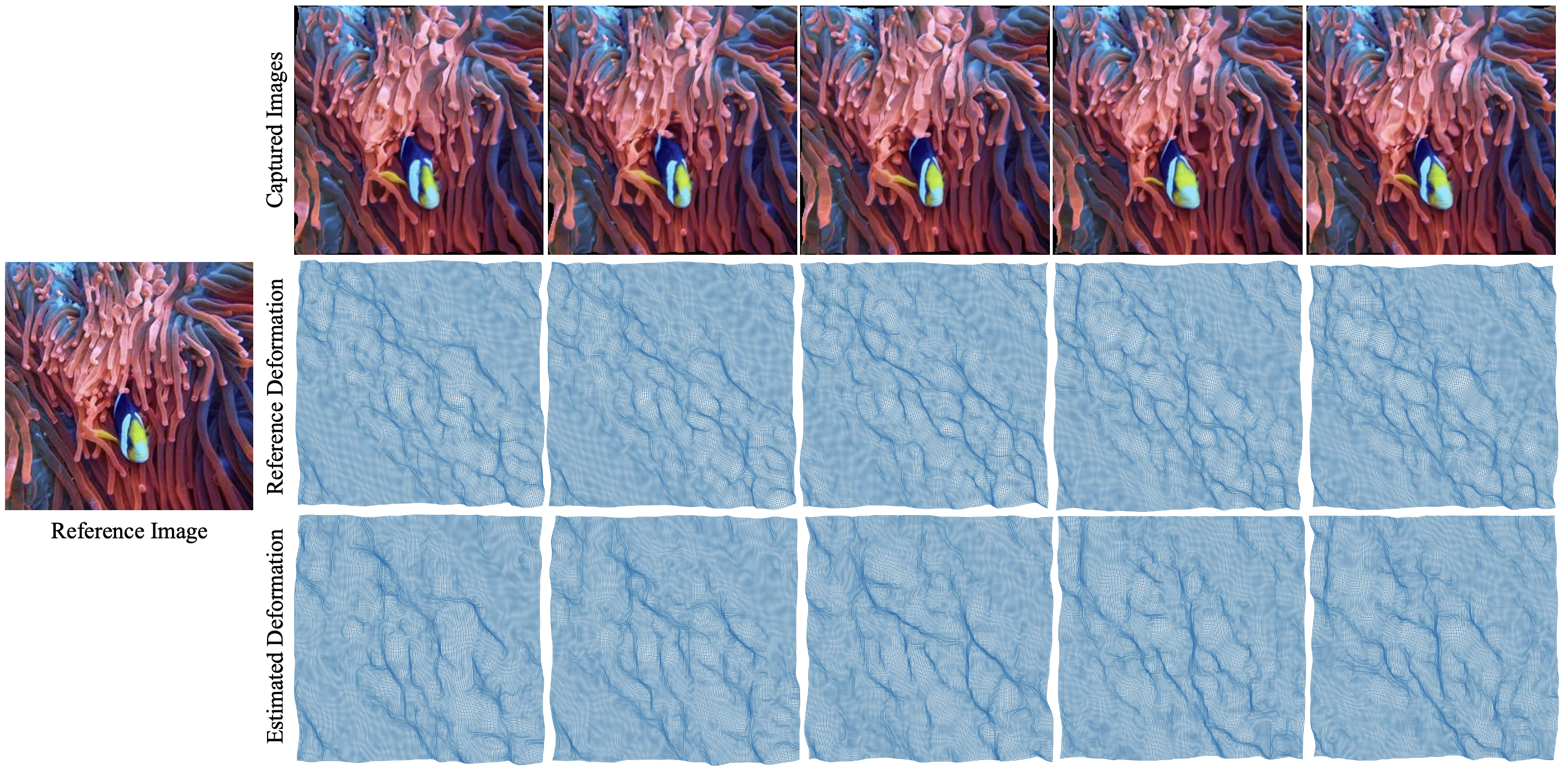}
    \includegraphics[width=0.83\linewidth]{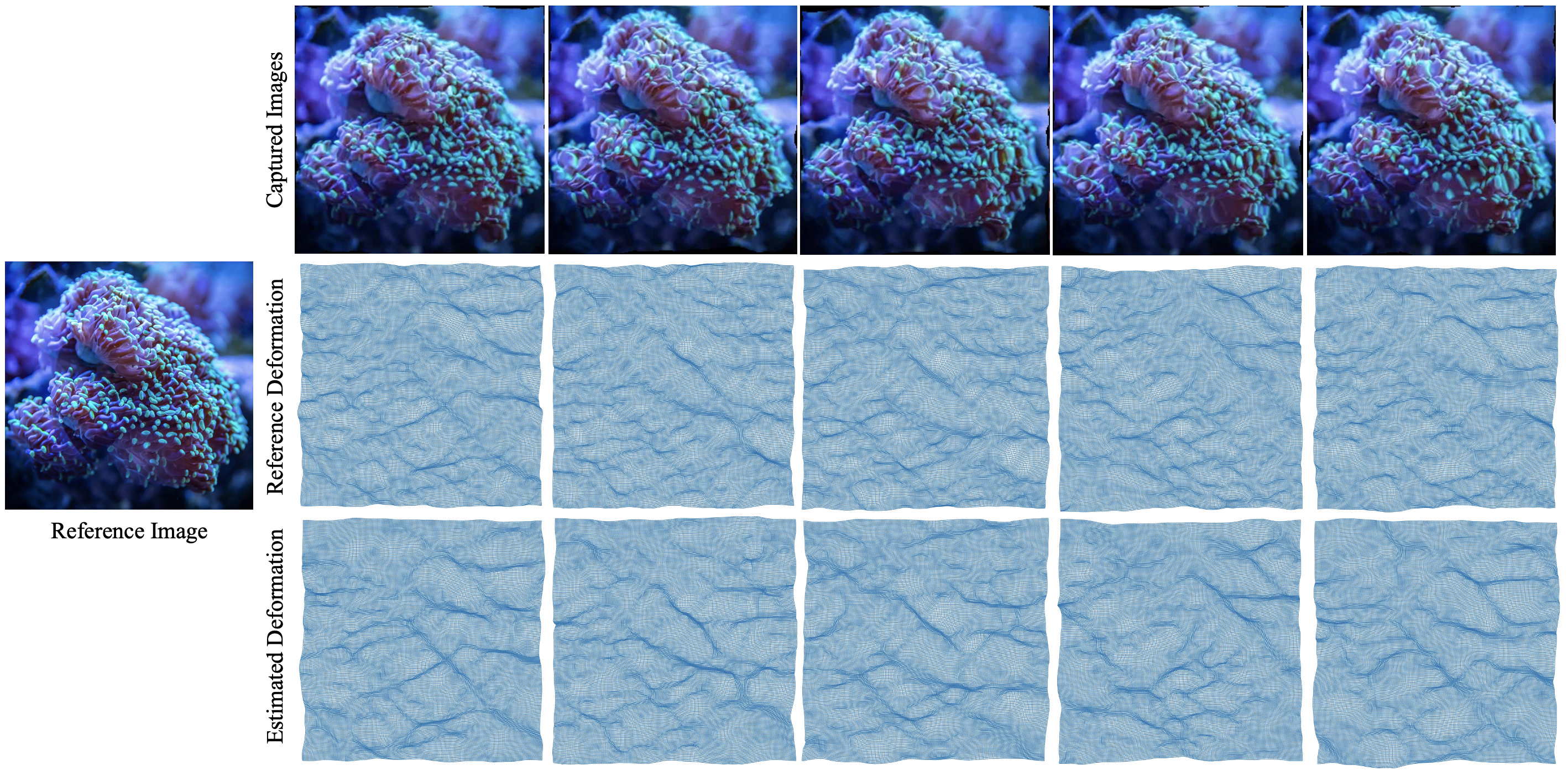}
    \includegraphics[width=0.83\linewidth]{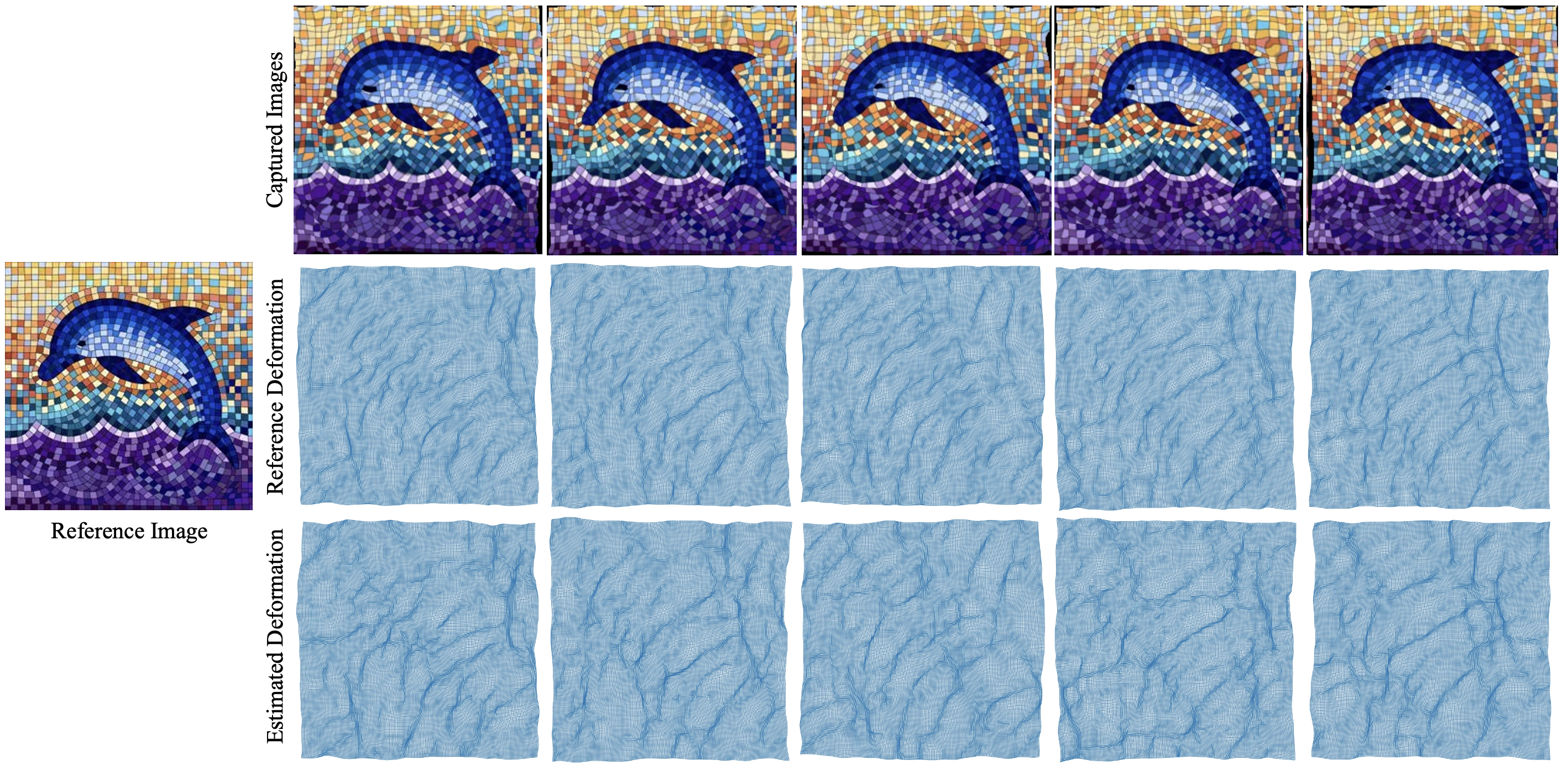}
    \caption{Evaluation on distortion field estimation under various turbulence intensities and directions. In each scenario, captured images (top) are sampled by applying reference deformations (middle) to the reference image (left). Our deformation estimations (bottom) are closely aligned with the reference fields. }
    \label{fig:deformations}
\end{figure*}

As shown in Figure~\ref{fig:deformations}, each example is presented with a high-quality reference image in the leftmost column, five successively distorted frames on the top row in each group, the corresponding five reference deformation fields used for simulation in the middle row, and the estimated deformation fields predicted by CQCD presented in the bottom row.

Across all cases, the estimated deformation fields exhibit a striking similarity to the ground-truth distortions. CQCD faithfully recovers the overall deformation structure — including orientation, density, and scale—suggesting its strong ability to disentangle and represent real turbulence-induced geometric distortions. Minor deviations occur only in flat-color regions, where the input images carry little visual information to infer the deformation.

The close match between estimated and reference deformation fields confirms that CQCD not only restores turbulence-corrupted images with high fidelity but also offers a physically interpretable representation of the underlying distortion process. This substantiates the theoretical soundness and practical robustness of our method in real-world water turbulence scenarios.


\subsection{Self Ablation}
\begin{table*}[!ht]
\small
\centering
\begin{tabular}{ccccccc}
\hline
\textbf{Turbulence} & \textbf{BC Weight} & \textbf{PSNR↑(dB)} & \textbf{SSIM↑} & \textbf{BC Loss} & \textbf{Invertible} \\
\hline
\multirow{4}{*}{Mild} & 0 & 20.40 & 0.7518 & 18.8477 & \ding{55} \\
& 0.01 & 21.60 & 0.7697 & 0.2895 & \checkmark \\
& 0.1 & 27.02 & \textbf{0.9097} & 0.0646 & \checkmark \\
& 1 & \textbf{27.27} & 0.8464 & 0.0106 & \checkmark \\
\hline
\multirow{4}{*}{Medium} & 0 & 15.63 & 0.5290 & Inf & \ding{55} \\
& 0.01 & 15.38 & 0.5571 & 0.4350 & \checkmark \\
& 0.1 & 18.74 & \textbf{0.8039} & 0.1326 & \checkmark \\
& 1 & \textbf{18.82} & 0.7618 & 0.0243 & \checkmark \\
\hline
\multirow{4}{*}{Severe} & 0 & 15.02 & 0.5058 & 16.3047 & \ding{55} \\
& 0.01 & 15.61 & 0.1974 & 6.0329 & \ding{55} \\
& 0.1 & \textbf{20.16} & \textbf{0.8212} & 0.0646 & \checkmark \\
& 1 & 19.18 & 0.7148 & 0.0106 & \checkmark \\
\hline
\end{tabular}
\caption{Impact of BC regularization weights on restoration performance. The best results are in bold.}
\label{tab:bc_ablation}
\end{table*}

\begin{table}[!ht]
\centering
\begin{tabular}{cccccc}
\hline
\textbf{Turbulence} & \textbf{Encoding} & \textbf{PSNR↑(dB)} & \textbf{SSIM↑} \\
\hline
\multirow{2}{*}{Mild} & w/ TF & \textbf{27.02} & \textbf{0.9097} \\
& w/o TF & 24.25 & 0.8335 \\
\hline
\multirow{2}{*}{Medium} & w/ TF & \textbf{18.74} & \textbf{0.8039} \\
& w/o TF & 17.03 & 0.6790 \\
\hline
\multirow{2}{*}{Severe} & w/ TF & \textbf{20.16} & \textbf{0.8212} \\
& w/o TF & 19.44 & 0.7448 \\
\hline
\end{tabular}
\caption{Impact of tight-frame (TF) feature extraction. The best results are in bold.}
\label{tab:tf_ablation}
\end{table}

\begin{table}[htbp]
\centering
\begin{tabular}{cccc}
\toprule
\textbf{Depth/Ratio} & \textbf{PSNR (dB)} & \textbf{SSIM} & \textbf{Time (s)} \\
\midrule
2 & 20.1607 & 0.8212 & 206 \\
3 & 20.1218 & 0.8250 & 241 \\
4 & 20.1635 & 0.8208 & 292 \\
5 & 20.0551 & 0.8200 & 324 \\
6 & 19.8443 & 0.8183 & 454 \\
\midrule
10\% & 19.8465 & 0.7371 & 166 \\
30\% & 20.0526 & 0.8149 & 181 \\
50\% & 20.1607 & 0.8212 & 206 \\
70\% & 19.9311 & 0.7913 & 261 \\
90\% & 19.7425 & 0.7773 & 290 \\
\bottomrule
\end{tabular}
\caption{Ablation studies on U-Net depth and training ratio of blur remover in every 100 iterations.}
\label{tab:ablation_studies}
\end{table}

\begin{figure*}
    \centering
    \includegraphics[width=\textwidth]{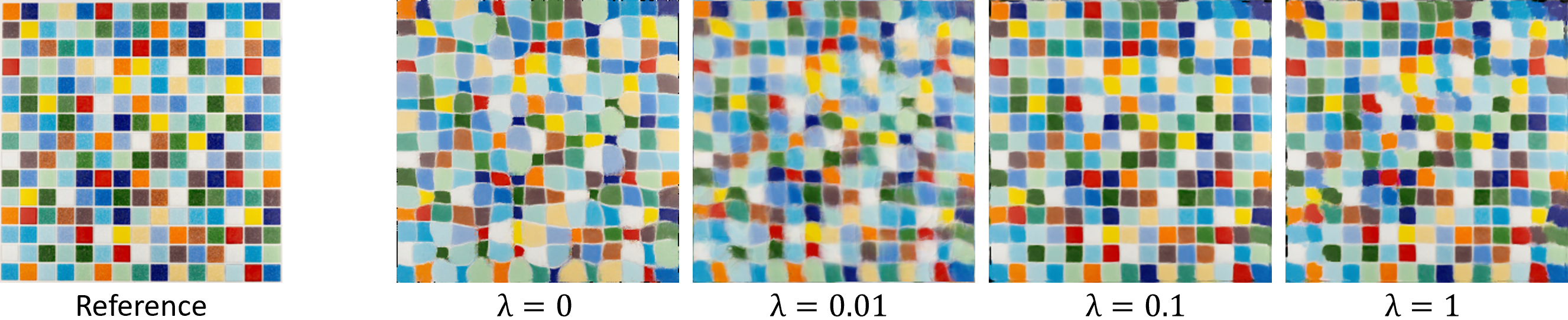}
    \caption{Visual comparison on deturbulence results by the influence of different BC weighting parameters in the proposed CQCD model.}
    \label{fig:selfablation}
\end{figure*}

\begin{figure*}
    \centering
    \includegraphics[width=0.6\linewidth]{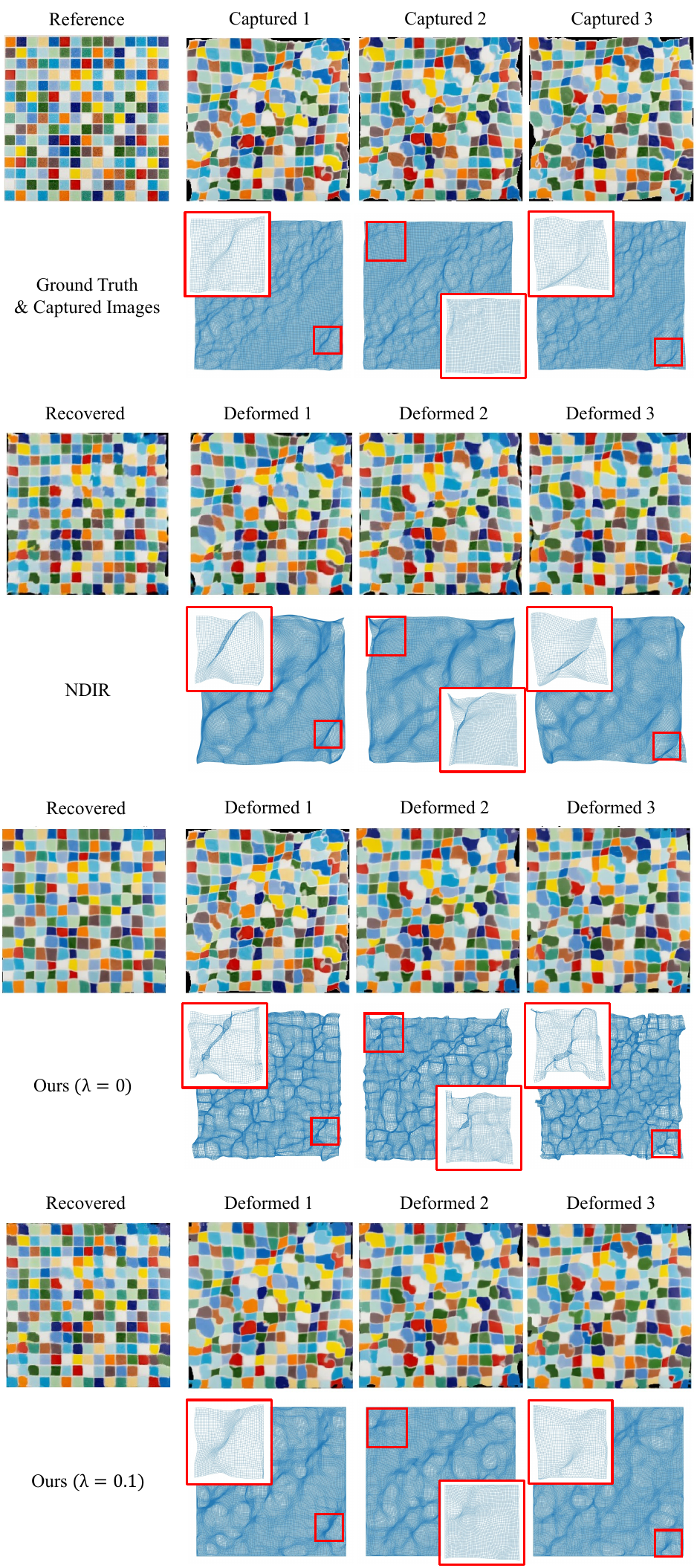}
    \caption{Study of quasi-conformal (QC) regularization for promoting} bijective mappings. The deformation grids are shown underneath their corresponding distorted images with local patches zoomed in. Visual comparison reveals that without QC constraints ($\lambda=0$), estimated deformation fields exhibit non-bijective folding (see grids), resulting in poor restoration and redistortion. This issue similarly affects the baseline NDIR. In contrast, our full model ($\lambda=0.1$) encourages invertibility via QC regularization, yielding both a clean restored image and more stable forward/inverse mappings that are essential for the circular architecture, especially when integrating multiple (potentially low‑quality) frames.
    \label{fig:bijectivity}
\end{figure*}

\begin{figure*}
    \centering
    \includegraphics[width=\textwidth]{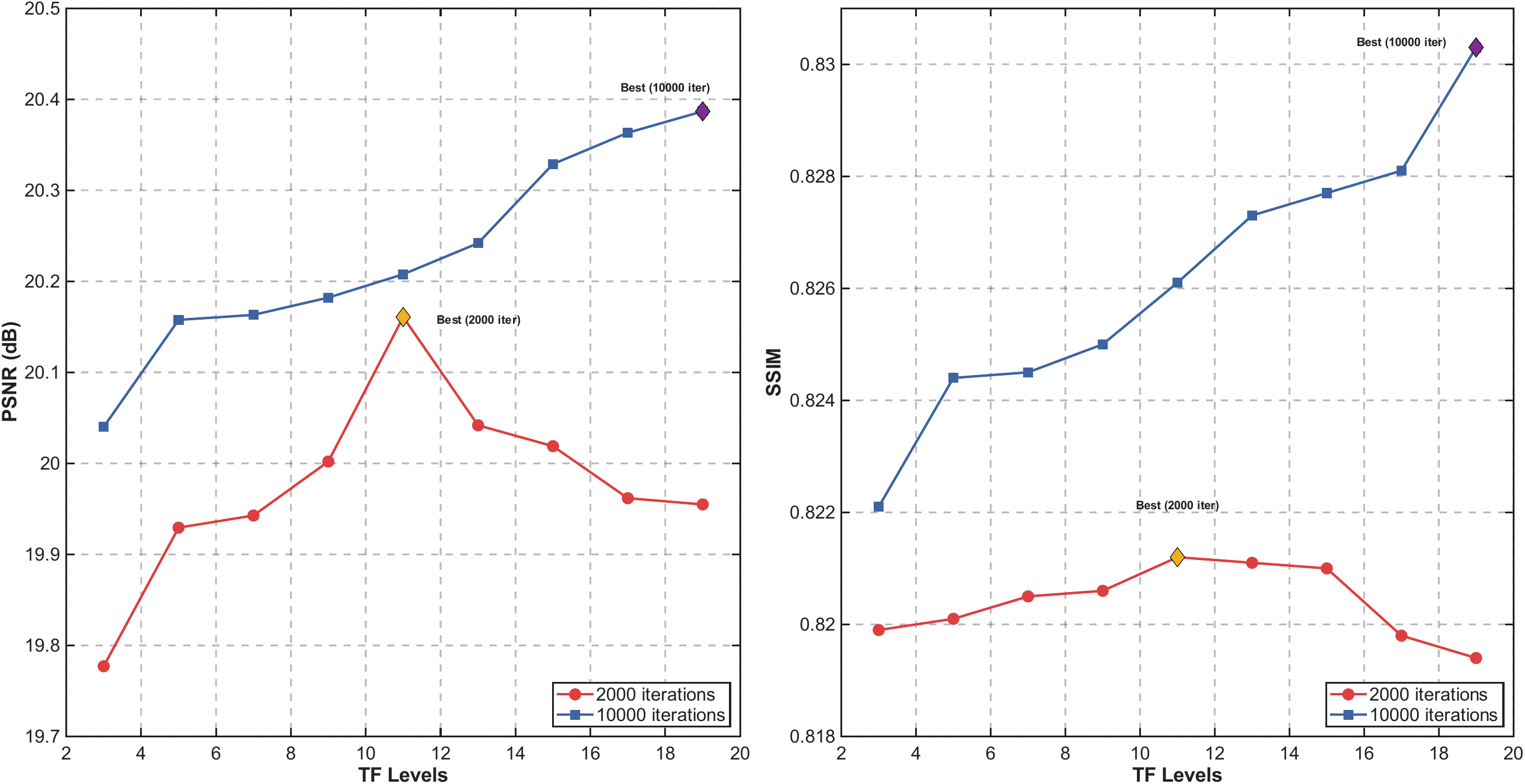}
    \caption{Ablation study on the number of TF decomposition levels. PSNR (left) and SSIM (right) vs. decomposition levels, respectively, under different training iterations. Optimal points for 2000 iterations and 10000 iterations are marked with diamonds. Results demonstrate that deeper decompositions capture finer directional details but require longer training to converge.}
    \label{fig:tf_levels}
\end{figure*}

The ablation studies underscore the critical contributions of QC regularization and TF encoding within the CQCD framework. For both ablation studies on BC weighting parameters and TF encoding, 10 frames are used for restoration.

\subsection{Test on QC Regularization}
\label{sec:qcregularization}
As shown in Table~\ref{tab:bc_ablation}, the choice of the BC weighting parameter $\lambda$ impacts restoration quality. When $\lambda=0$ (no QC regularization), the framework is likely to naively fit into a local optimum, minimizing the reconstruction loss but failing to correct distortions, as evidenced by residual warping in Figure~\ref{fig:selfablation}. With $\lambda=0.01$, insufficient BC constraints lead to unstable deformation adjustments, where the estimated fields oscillate excessively during training, causing inconsistent blur removal and incomplete distortion recovery. In both cases, the lack of sufficient BC weighting results in a non-invertible mapping, making it impossible to maintain the circular consistency required by the framework. Notably, color bleeding effects are observed due to the absence of orientation-preserving enforcement provided by QC regularization.

On the other side, when $\lambda = 1$, although the framework still performs reasonably under mild and medium turbulence, restoration quality degrades under severe turbulence. In this case, the overly rigid QC regularization excessively constrains deformation flexibility, leading to undercorrected geometric distortions and potentially requiring more computation time to reach good results. In contrast, setting $\lambda = 0.1$ strikes an effective balance: the moderate QC constraint encourages physically plausible and practically invertible deformations while retaining enough flexibility to adapt to local turbulence. This setting yields the best overall performance across all distortion levels.

The robustness of CQCD stems from the synergistic combination of the circular architecture with the QC regularization. While the circular design enforces self‑consistency across frames, it critically depends on the estimated deformation fields being practically invertible. Without explicit geometric constraints, as in NDIR or our ablated model without QC ($\lambda=0$), the estimated mappings often become non‑bijective, exhibiting self‑folding and discontinuities (Figure~\ref{fig:bijectivity}). Such mappings are difficult to invert stably, breaking the cycle‑consistency loop and causing the model to accumulate errors when more frames are added. In contrast, the QC regularization encourages smoother and more bijective deformation fields, improving the well-definedness of forward and inverse transformations in practice. This geometric stability allows the circular framework to integrate information from multiple frames—even those with strong distortions—in a coherent manner, preventing the performance degradation observed in methods that lack such constraints. Thus, the QC regularization is not merely an auxiliary loss: it empirically reduces foldings and helps the circular architecture function reliably as the number of input frames increases.

\subsection{Test on Tight-Frame Encoding}

The necessity of TF encoding is further highlighted in Table~\ref{tab:tf_ablation}. Removing the TF block degrades performance across all turbulence levels. This decline stems from the loss of directional sensitivity: the ability of the TF transform to extract orientation-aware, multi-scale features is essential for disentangling coherent turbulence-induced distortions such as ripples and elongated warping. Without this encoding, the model struggles to maintain stable and discriminative representations, particularly under complex turbulence. Furthermore, the ablation study (Figure~\ref{fig:tf_levels}) on TF decomposition levels reveals a fundamental trade‑off between representational capacity and optimization efficiency. When trained with a fixed computational budget (2000 iterations), moderate decomposition depth (11 levels) yields optimal performance, balancing the ability to capture multi‑scale directional features with stable convergence. This configuration provides sufficient hierarchical representation to disentangle turbulence‑induced distortions at multiple frequencies while avoiding the under‑fitting that occurs with shallower decompositions or the over‑parameterization that impedes convergence within limited training. However, when the iteration is increased (10000 iters), deeper decompositions (up to 19 levels) progressively improve restoration quality, demonstrating that additional levels can model finer directional details and more complex geometric warping patterns, provided sufficient optimization time is allocated. This behavior confirms the theoretical advantage of multi‑level TF transforms: each additional level extracts increasingly fine‑grained, orientation‑aware features that are essential for accurately representing the anisotropic, multi‑scale nature of turbulence distortions. In practice, the choice of 11 levels represents a pragmatic balance, delivering robust performance within reasonable training constraints while acknowledging that further gains are achievable with extended optimization. This finding underscores the importance of matching the model’s representational depth with the available computational resources, and it validates the inclusion of multi‑scale directional feature extraction as a core component of our framework.

\subsection{Test on U-Net Depth and Alternating Ratio}

Additionally, the ablation studies on U-Net depth and the training ratio between the blur remover and mapping estimator further validate our architectural and optimization choices. As shown in Table~\ref{tab:ablation_studies}, varying the depth of the deformation estimator reveals that a shallow U-Net with depth 2 already achieves competitive performance while requiring the shortest training time (206s). Increasing the depth to 4 yields a marginal PSNR gain at the cost of $42\%$ longer training, whereas deeper architectures exhibit clear degradation in both metrics alongside substantially increased runtime. This indicates that the deformation estimation task in our framework does not benefit from excessive network capacity, a relatively compact U-Net suffices to capture the quasi‑conformal deformation fields, and deeper networks may over‑fit or complicate the optimization within the circular loop. Regarding the training ratio, a balanced allocation (50-50 in every 100 iterations) between the blur remover and the mapping estimator delivers the best performance. Ratios skewed toward either network (0.1 or 0.9) lead to noticeable performance drops, suggesting that both components require adequate updates to maintain the synergistic correction of geometric and photometric distortions. Consequently, the combination of depth 2 and ratio 0.5 represents an effective trade‑off that preserves restoration quality while minimizing computational overhead, and we adopt this configuration as the default in all other experiments.
\section{Conclusion}
In this paper, we propose Circular Quasi-Conformal Deturbulence (CQCD), a novel unsupervised framework for restoring images degraded by complex geometric distortions, such as those caused by inhomogeneous media. CQCD restores images by leveraging a circular self-consistency constraint that enforces both spatial and temporal coherence.

The model operates in a three-stage cycle: (1) estimating a deformation field to correct geometric turbulence, (2) deblurring the warped image, and (3) reproducing the original distortion using the inverse of the estimated deformation. To ensure physically plausible mappings, the deformation is regularized via quasi-conformal (QC) constraints, while tight-frame (TF) transformations are used for multi-scale, direction-aware feature extraction.

We validate the effectiveness of CQCD through extensive experiments on both synthetic turbulence sequences (with varying frame counts) and real distorted images, benchmarking against state-of-the-art methods. We further evaluate the accuracy of the estimated deformation fields by comparing them with ground-truth distortions used in simulation. Finally, self-ablation studies on the QC regularization weight and the inclusion of TF blocks confirm their critical roles in achieving stable and accurate restoration.

\section{Limitation and Future Work}

A fundamental limitation of the proposed framework is that it is designed for static scenes. Within the circular quasi-conformal (CQCD) architecture, two key optimization components rely on pixel position–based discrepancies: (i) minimizing the difference between the estimated distorted image and the captured image, and (ii) enforcing similarity consistency over resampled images. Both objectives implicitly assume pixel-wise correspondence across images. When objects within the scene move, or when global camera motion is present, this assumption is violated. In such cases, temporal changes are difficult to reconcile using position-based pixel differences, and the model may incorrectly interpret motion-induced inconsistencies as geometric distortion. 

Additionally, since quasi-conformal regularization is a key component of our circular framework that ensures invertibility of the distortion mapping, the model is theoretically limited to bijective mappings. Consequently, scenes that involve topology-changing geometric deformations fall outside the scope of our approach, and the current model is not suitable for handling such distortions.

In future work, extending the CQCD framework to dynamic scenes with moving objects or camera motion remains a challenging but natural direction. Addressing this limitation would require more robust motion disentanglement and correspondence alignment mechanisms, such as joint deformation–motion modeling or temporally correspondence matching, to reliably separate underlying geometric distortions from genuine object or scene dynamics. Furthermore, we plan to explore additional feature extraction methods, such as Wavelet Transformations in the frequency domain \cite{shang2025training,shang2025waveletmamba}, which can effectively decouple high- and low-frequency components. These features are promising for improving the model’s learning capability and should complement the existing tight-frame features.

\section*{Acknowledgments}
This work was supported by HKRGC GRF (Project IDs: 14306721 and 14307622), and Hong Kong Centre for Cerebro- Cardiovascular Health Engineering (COCHE).

\begin{scriptsize}
\bibliographystyle{unsrt}  
\bibliography{reference}
\end{scriptsize}

\end{document}